\documentclass[letterpaper]{article} 
\usepackage[preprint]{aaai2027}  
\usepackage[hyphens]{url}  
\usepackage{graphicx} 
\usepackage{natbib}  
\usepackage{caption} 
\usepackage{algorithm}
\usepackage{algorithmic}

\usepackage{kotex}
\usepackage{multirow}
\usepackage{subcaption}
\usepackage{newfloat}
\usepackage{listings}
\DeclareCaptionStyle{ruled}{labelfont=normalfont,labelsep=colon,strut=off} 
\floatstyle{ruled}
\newfloat{listing}{tb}{lst}{}
\floatname{listing}{Listing}

\usepackage{booktabs}

\usepackage{enumitem}
\usepackage{amsmath}
\usepackage{amssymb}

\newif\ifshownotes \shownotestrue

\newcommand{\R}{\mathbb{R}}
\newcommand{\C}{\mathbb{C}}
\newcommand{\Fop}{\mathcal{F}}
\newcommand{\Sone}{\mathbb{S}^{1}}
\newcommand{\GL}{\mathrm{GL}}
\newcommand{\E}{\mathbb{E}}
\newcommand{\src}{\mathrm{s}}
\newcommand{\tgt}{\mathrm{t}}
\newcommand{\Xs}{X^{(\src)}}
\newcommand{\Xt}{X^{(\tgt)}}
\newcommand{\ws}{w^{(\src)}}
\newcommand{\wt}{w^{(\tgt)}}

\newcommand{\Lmag}{\mathcal{L}_{\text{mag}}}
\newcommand{\Lphasevel}{\mathcal{L}^{\text{vel}}_{\text{phase}}}
\newcommand{\Lphaseq}{\mathcal{L}^{\text{q}}_{\text{phase}}}
\newcommand{\Lipd}{\mathcal{L}_{\text{IPD}}}
\newcommand{\Lrec}{\mathcal{L}_{\text{rec}}}

\title{RIPPLE: Generating Multi-Channel Phase, Not Recovering It}
\author{
    Jaehyuk Lee\textsuperscript{\rm 1},
    Yeajin Lee\textsuperscript{\rm 2},
    Dayeon Shin\textsuperscript{\rm 1},
    Donghun Lee\textsuperscript{\rm 1}\corresponding
}
\affiliations{

    \textsuperscript{\rm 1}Department of Mathematics, Korea University\\
    \textsuperscript{\rm 2}Program in Actuarial Science and Financial Engineering, Korea University\\
    145, Anam-ro, Seongbuk-gu, Seoul, Republic of Korea\\
    \{jaehyeokbear, gin1214, yeon\_25, holy\}@korea.ac.kr
}

\begin{document}

\maketitle

\begin{abstract}
Generative models synthesize magnitude spectra with high fidelity, while phase is delegated to a recovery module---Griffin--Lim, a vocoder, or a latent decoder---applied independently to each channel. For multi-channel waveforms this delegation is costly: the physical content of spatial audio and three-component seismograms lives in the phase relationships between channels, precisely what channel-independent recovery cannot produce. The cost is also invisible, since the magnitude-based metrics common to both fields barely move when inter-channel phase coherence collapses---so a pipeline can discard the physical information in its output while still scoring well. We argue that phase should be generated, not recovered, and present RIPPLE (Rectified Inter-channel Phase with Prior-based LEarning), which reinterprets Griffin--Lim as a phase \emph{prior} rather than a final estimator: initialized from the source phase, this prior carries the inter-channel structure to be preserved, and a rectified flow refines it toward the target under an explicit inter-channel phase loss. Tested on first-order ambisonics environment transfer and seismic cross-station translation---two physically unrelated domains---RIPPLE outperforms recovery-based pipelines on the coherence metrics that downstream analyses consume. The seismic case is decisive: across architecturally distinct generators, per-channel recovery leaves S-wave polarization error near the $57.3^\circ$ random expectation, whereas learned phase reduces it to $33.8^\circ$.
\end{abstract}

\section{Introduction}
Generative modeling of waveforms has advanced largely by splitting the short-time Fourier transform (STFT) into a magnitude that is generated and a phase that is \emph{reconstructed}.
This division has been productive for single-channel signals, where a magnitude-consistent phase estimate usually suffices, so phase has been handed off to separate recovery or decoding modules.

The assumption no longer holds for \emph{multi-channel} waveforms, for which the \emph{inter-channel} relationship itself carries the information of interest.
In first-order ambisonics (FOA), source azimuth and elevation are encoded in the relative phase and amplitude of the W/Y/Z/X channels; in three-component (Z/N/E) seismograms, P- and S-wave arrival times and particle motion are jointly determined by the relative amplitudes and the coherent phase structure across components. Independent per-channel phase recovery does not merely degrade inter-channel information; it can eliminate it altogether.

This limitation is not specific to any one recovery mechanism. 
We measure the routes both fields use under oracle conditions: Griffin–Lim on ground-truth magnitudes, a strong neural vocoder (BigVGAN), a spatial codec round-trip, and a seismic latent-VAE round-trip. All of them fail in the same way. In spatial audio, azimuth error reaches the chance level ($\pi/2$); in seismic data, S-wave polarization error increases from $3.7^\circ \to 42.9^\circ$, whereas amplitude-based GOF remains high. These errors arise with no generative model in the loop (Table~\ref{tab:oracle}). Post-hoc phase recovery imposes a ceiling that no generator built on it can exceed.

Two facts let this failure ship unnoticed. The consumer of these waveforms is not a listener but an algorithm---a localizer reading direction from a re-rendered intensity vector, or a polarization analysis reading particle motion---so it consumes exactly the inter-channel phase that per-channel recovery destroys; and that destruction is silent, since the waveform-, spectral-, and magnitude-based goodness-of-fit metrics both fields default to barely move when inter-channel phase lands at or near chance.

Learning phase directly is awkward: it wraps around on $\Sone$ and is entangled multiplicatively with magnitude. The common response is to avoid learning it at all.
We take the opposite position: for multi-channel waveforms, phase is not a nuisance variable to be recovered but a structured target to be generated, and the structure classical estimators capture should enter as a prior, not as the final answer.

\paragraph{What existing approaches miss.}
According to our survey, summarized in Table~\ref{tab:novelty_main}, no prior work pairs learned phase with source coherence, a cross-channel objective, and coherence-sensitive evaluation. 
In other words, phase is handled without an inter-channel objective, kept outside the generative model---delegated to per-channel recovery or a compressed latent---or generated per channel with no coherence target. 
Such approaches cannot extend well to FOA; the authors of Diff-SAGe make the same argument, and tries to remedy it by modeling phase explicitly as a complex spectrogram \citep{kushwaha2025diffsage}. 

We take that argument one step further and make inter-channel coherence itself the training target.
At the same time, we observe that representing phase explicitly does not by itself produce phase that is coherent across channels. When we adapt Diff-SAGe to our source-conditioned setting, its phase fidelity degrades rather than transfers (Table~\ref{tab:spatial_main}). Seismology fits the same pattern, with phase recovered by per-channel Griffin–Lim \citep{jung2025heggs} or test-time retrieval \citep{bi2025advancing}.

\paragraph{Why translation, and why a prior.}
When two waveforms from the same source can be paired, such that the same underlying signal is realized through two transfer functions (e.g. two room impulse responses, two geological paths), phase generation can be reframed as translation from a source to a target. 
Unlike synthesis from mono or semantic input, which must conjure coherence from a signal that has none, translation need only transport it. We use Griffin--Lim as phase prior to seed this structure, turning phase generation into a correction problem (quantified in our ablations).

\textbf{Contributions.}
We present \textbf{RIPPLE} (\textbf{R}ectified \textbf{I}nter-channel \textbf{P}hase with \textbf{P}rior-based \textbf{LE}arning) contributing: 
\begin{itemize}[leftmargin=*,itemsep=1pt,topsep=2pt]
\item \textbf{Utilizing Griffin--Lim as a phase prior, not a phase decoder.}
Initialized from the source phase, it supplies magnitude consistency without destroying source coherence, reducing phase generation to a residual correction.

\item \textbf{Decoupled two-stage rectified flow with an inter-channel phase-difference (IPD) objective.}
Independent timesteps $(t_m, t_p)$ decorrelate the two modalities' learning signals, and the inter-channel phase-difference loss makes coherence an explicit training target.

\item \textbf{Cross-domain empirical validation.} 
On FOA environment transfer and on seismic cross-station translation---a physically unrelated domain---RIPPLE outperforms recovery- and vocoder-based pipelines on coherence metrics, most sharply on S-wave polarization.
\end{itemize}

\section{Related Work}

\paragraph{Multi-channel spatial audio generation.}
Neural multi-channel synthesis spans mono-to-binaural rendering \citep{richard2021warpnet, leng2022binauralgrad, lee2023nfs, liang2025binauralflow}, FOA generation via spatial codecs \citep{heydari2025immersediffusion} or complex-spectrogram flow transformers \citep{kushwaha2025diffsage}, and video- or language-conditioned models \citep{liu2025omniaudio, templin2025sonicmotion}, all mapping mono or semantic input, which carries no inter-channel phase structure, to spatial output. 

\paragraph{Seismic waveform and ground-motion generation.}
Three-component waveforms have been synthesized with conditional GANs \citep{wang2021seismogen, li2024conseisgen} and latent diffusion on spectrograms \citep{bergmeister2025highfem, bi2025advancing}. In all of these, phase is recovered rather than learned—via test-time phase retrieval \citep{bi2025advancing} or per-channel Griffin–Lim as in HEGGS \citep{jung2025heggs}; we adopt the latter as our directly comparable baseline to isolate whether learning multi-component phase improves inter-component coherence.

\begin{table}[t]
    \centering
    \caption{Positioning of the closest surveyed work.
    \emph{Rec}: phase delegated to recovery; \emph{impl.}: waveform
    generation without separate phase; \emph{Gen}: phase as explicit
    target. ``--'': not established by the cited source;
    $\triangle$: partial. Full version with representations and
    I/O settings in Appendix~A.}
    \label{tab:novelty_main}
    \footnotesize
    \setlength{\tabcolsep}{2pt}
    \renewcommand{\arraystretch}{1.1}
    \begin{tabular}{@{}lcccc@{}}
    \toprule
    \textbf{Method} & \textbf{Phase} & \textbf{Src.\ coh.} & \textbf{X-loss} & \textbf{Coh.\ eval} \\
    \midrule
    Mono/video/lang.\ synth.\textsuperscript{1} & impl. & $\times$ & $\times$ & -- \\
    BinauralFlow\textsuperscript{2}    & Gen & $\times$ & $\times$ & $\triangle$ \\
    Tango+voc.\ / ImmerseDiff.\textsuperscript{3} & Rec & $\times$ & $\times$ & $\times$ \\
    Diff-SAGe\textsuperscript{4}       & Gen & $\times$ & $\times$ & $\triangle$ \\
    FOA Tokenizer\textsuperscript{5}   & Rec & n/a & $\triangle$ & -- \\
    GANs / latent diffusion\textsuperscript{6} & impl./Rec & $\times$ & $\times$ & -- \\
    HEGGS\textsuperscript{7}           & Rec & \checkmark & $\times$ & $\times$ \\
    \textbf{RIPPLE} (ours) & \textbf{Gen} & \checkmark & \checkmark & \checkmark \\
    \bottomrule
    \end{tabular}
    
    \vspace{2pt}
    {\scriptsize\raggedright
    \textsuperscript{1}\citet{richard2021warpnet,leng2022binauralgrad,lee2023nfs,liu2025omniaudio,templin2025sonicmotion};
    \textsuperscript{2}\citet{liang2025binauralflow};
    \textsuperscript{3}\citet{ghosal2023text,lee2022bigvgan,heydari2025immersediffusion};
    \textsuperscript{4}\citet{kushwaha2025diffsage};
    \textsuperscript{5}\citet{sudarsanam2025foatokenizer};
    \textsuperscript{6}\citet{wang2021seismogen,li2024conseisgen,bergmeister2025highfem,bi2025advancing};
    \textsuperscript{7}\citet{jung2025heggs}.\par}
\end{table}

\paragraph{Phase recovery and neural vocoders.}
The Griffin--Lim algorithm \citep{griffin1984} estimates a magnitude-consistent phase by alternating projections. Neural vocoders replace this with learned synthesis from magnitude-like features---adversarially trained waveform generators such as HiFi-GAN and BigVGAN \citep{kong2020hifigan, lee2022bigvgan}, or Fourier-coefficient prediction as in Vocos \citep{siuzdak2024vocos}---but operate per signal, without a cross-channel objective; FOA Tokenizer \citep{sudarsanam2025foatokenizer} adds a spatial-consistency loss, though for compression rather than generation.

\paragraph{Flow matching and rectified flows for audio.}
Flow matching \citep{lipman2023flow}, stochastic interpolants \citep{albergo2023interpolants}, and rectified flow \citep{liu2023rectified} are now widely used for audio \citep{guan2024lafma, liu2025omniaudio, liang2025binauralflow}; manifold-native variants instead define the interpolant on the geometry itself \citep{mathieu2020riemannian, chen2024flow}, an alternative we discuss in Appendix~\ref{app:p_0}.
Diffusion-bridge methods \citep{liu2023i2sb, ICLR2024_20e45668} likewise start from an informed endpoint rather than noise, and multi-modal diffusion has used per-modality timesteps \citep{bao2023one}. Our contribution is not the transport machinery but where transport starts from: a magnitude-consistent, source-coherent Griffin--Lim prior.

\textbf{Positioning.}
Table~\ref{tab:novelty_main} makes the comparison explicit: within the surveyed work, RIPPLE is the only method that pairs learned phase with source coherence, a cross-channel objective, and coherence-sensitive evaluation (full version in Appendix~\ref{app:positioning}).

\section{Method}

\begin{figure*}[t]
    \centering
        \includegraphics[width=0.8\linewidth]{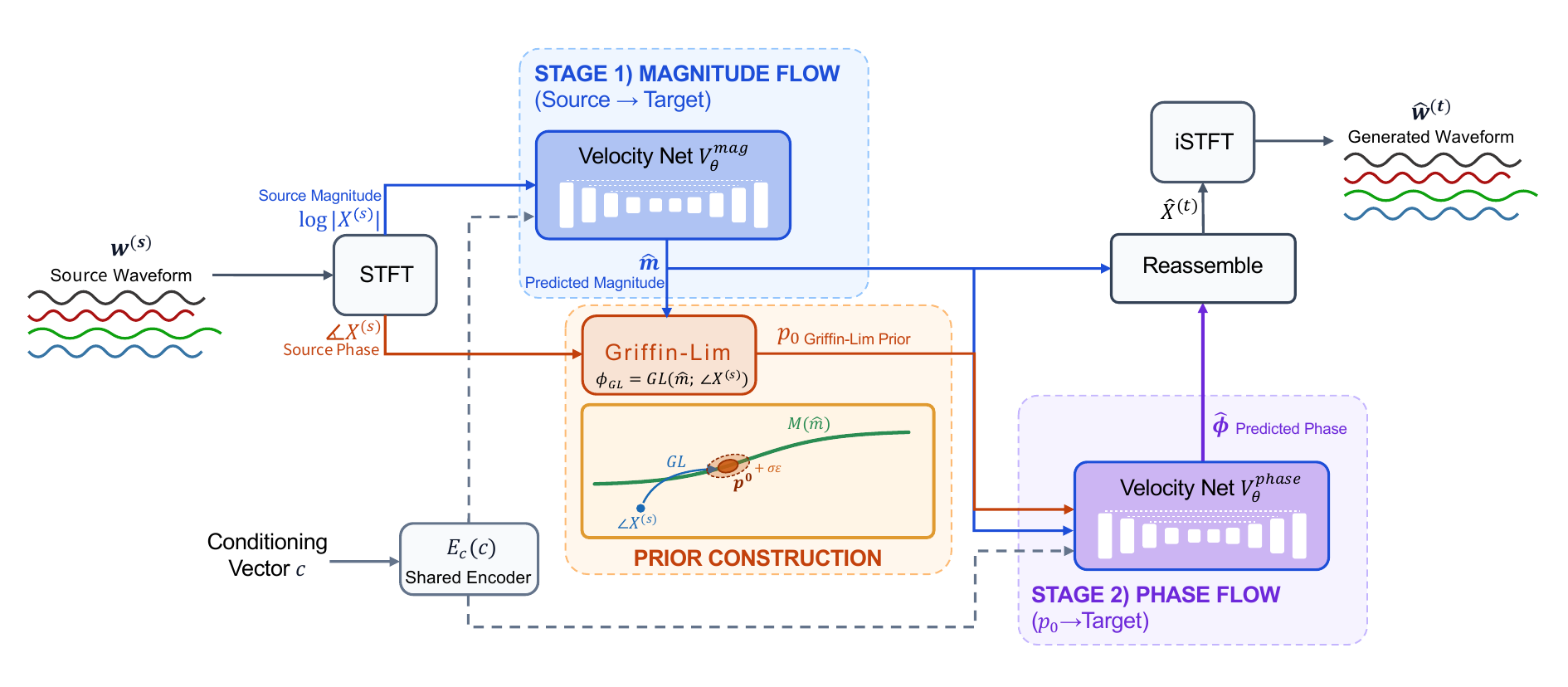}
        \caption{Overview of RIPPLE. Stage 1 predicts the target magnitude; the source-informed Griffin–Lim prior is built from it; Stage 2 integrates the phase flow from this prior.}
        \label{fig:main}
\end{figure*}

\subsection{Problem Setup}
We study source-to-target translation of multi-channel waveforms.
Given a $C$-channel source $\ws\in\R^{C\times L}$ and a target $\wt\in\R^{C\times L}$ that is the \emph{same} underlying source signal passed through a different transfer function, together with a target conditioning vector $c\in\R^{d_c}$ (position / room / distance metadata), we learn the conditional distribution $p(\wt \mid \ws, c)$.
Our two instantiations are FOA spatial audio ($C{=}4$) and three-component seismograms ($C{=}3$); see Experimental Setup in Appendix~\ref{app:exp_details}.

\subsection{Spectrogram Representation}
Let $X=\Fop\{w\}\in\C^{C\times F\times T}$ be the complex STFT.
We represent it in the disentangled form
\begin{equation}
Z = \big(\log|X|,\ \cos\angle X,\ \sin\angle X\big)\in\R^{3C\times F\times T}.
\end{equation}
Two design choices motivate this.
\textbf{(a) $\Sone$ phase embedding.} Phase $\phi\in\Sone$ is embedded into $\R^2$ as $(\cos\phi,\sin\phi)$, removing the wrap-around discontinuity by construction: 
naive regression penalizes the identical points $\phi{=}\pm\pi$ by $(2\pi)^2{\approx}39$.
\textbf{(b) Magnitude/phase disentanglement.} 
Keeping magnitude and phase as separate fields, rather than folding them into $(\Re X,\Im X)$, lets magnitude serve as conditioning information for phase—the property our Griffin--Lim prior and magnitude-weighted phase loss both exploit—and lets the two modalities carry different priors.

\subsection{Griffin--Lim as a Magnitude-Consistent Phase Prior}
The Griffin--Lim algorithm $\mathrm{GL}_K(m;\phi_{\text{init}})$ seeks, by alternating projections, a phase $\phi$ such that $m\odot e^{i\phi}$ is a consistent STFT with magnitude m (i.e., a fixed point of $\Fop\{\Fop^{-1}\{\cdot\}\}$), iterating for $k=0,\dots,K{-}1$:
\begin{equation}
\phi^{(k+1)} = \angle\,\Fop\!\big\{\Fop^{-1}\{m\odot e^{i\phi^{(k)}}\}\big\},
\qquad \phi^{(0)}=\phi_{\text{init}}.
\end{equation}
Standard usage treats $\phi^{(K)}$ as the final phase estimate.
We instead treat it as an \emph{informed prior} for flow matching.
The iteration is applied independently per channel, so Griffin--Lim itself cannot generate inter-channel phase relationships; whatever coherence the prior carries must come from its initialization $\phi_{\text{init}}$.

\paragraph{Source-informed initialization.}
We initialize from the source phase, $\phi_{\text{init}}=\angle\Xs$.
Because the target and source share an underlying source signal, the source phase carries inter-channel structure 
that is correlated with the target's, though the two transfer functions generally reshape it per frequency.
Per-channel Griffin–Lim iterations gradually attenuate this structure. We add a small Gaussian perturbation to $p_0$, giving the perturbed prior of Eq. ~\ref{eq:p0}:

\begin{equation}\label{eq:p0}
    \begin{aligned}
        p_0 &= \big(\cos\phi_{\GL},\,\sin\phi_{\GL}\big)+\sigma_{\text{prior}}\,\epsilon, \\
        \phi_{\GL} &= \GL_K\!\big(m^{(\tgt)};\,\angle\Xs\big),
        \quad \epsilon\sim\mathcal{N}(0,I).
    \end{aligned}
\end{equation}

The perturbation trains the velocity field on a neighborhood of the prior rather than the exact prior alone---inference builds 
$p_0$ from the predicted $\hat{m}$, not the oracle $m^{(t)}$ used in training---and takes $p_0$ off the unit circle, supplying the radial support that integration in $\mathbb{R}^2$ requires (App.~\ref{app:p_0}).

\subsection{Decoupled Two-Stage Rectified Flow}
We model magnitude and phase with two rectified flows that use independent interpolation timesteps (Fig.~\ref{fig:main}).

\paragraph{Magnitude flow (source $\rightarrow$ target).}
\begin{align}
z^{\text{mag}}_{t_m} &= (1-t_m)\log|\Xs| + t_m\log|\Xt|,\\
v^{*}_{\text{mag}} &= \log|\Xt| - \log|\Xs|.
\end{align}

\paragraph{Phase flow (Griffin--Lim prior $\rightarrow$ target).}
\begin{equation}\label{eq:phase_inter}
    \begin{aligned}
    z^{\text{phase}}_{t_p} &= (1-t_p)\,p_0 + t_p\big(\cos\angle\Xt,\sin\angle\Xt\big),\\
    v^{*}_{\text{phase}} &= \big(\cos\angle\Xt,\sin\angle\Xt\big) - p_0.
    \end{aligned}
\end{equation}
The timesteps are sampled independently, $t_m,t_p\sim\mathcal{U}(0,1)$.
We use two separate velocity networks $v^{\text{mag}}_\theta$ and $v^{\text{phase}}_\theta$ that share the same conditioning encoder $E_c(c)$.
Since magnitude is informative for phase, the input to $v^{\text{phase}}_\theta$ additionally concatenates $\log|\Xt|$ along the channel dimension, making the target magnitude available to the phase head throughout integration.

\textbf{Why decoupled timesteps.} 
Since the two velocity networks share $E_c$, tying $t_m{=}t_p$ correlates their training noise levels; independent sampling decorrelates them, and the effect concentrates on inter-channel coherence (Table~\ref{tab:ablation_merged}).

\textbf{Why a Griffin--Lim prior helps.}
Starting from $p_0$ rather than noise places the phase flow near the STFT-consistency manifold at the target magnitude, so it learns only a residual correction—shortening transport and letting short schedules suffice (Figure~\ref{fig:nfe_curve}).

\subsection{Training Objective}\label{sec:loss}

The overall training objective is a weighted sum of two velocity-regression losses, a magnitude-reconstruction loss, phase-quality terms, and an inter-channel phase-difference (IPD) loss. All phase-related losses are magnitude-weighted so that low-energy bins, where phase is informationally degenerate, do not dominate training.
For convenience, we define the one-step prediction of each modality $\bullet \in \{\mathrm{mag}, \mathrm{phase}\}$ as

\begin{equation}
\hat z^{\bullet}_{1|t} = z^{\bullet}_t + (1-t)\, v^{\bullet}_\theta(\cdot),
\end{equation}

\paragraph{Magnitude-aware weighting.}
We weight phase losses by a per-sample normalized sigmoid gate
\begin{equation}\label{eq:mag_weight}
    w(|X^{(t)}|) = \sigma\!\left(\frac{\log|X^{(t)}| - \mu}{s}\right),
\end{equation}
where $\mu$ and $s$ are the per-channel mean and standard deviation of $\log|X^{(t)}|$ over the $(f, t)$ axes of each sample, and $w$ is subsequently normalized to unit mean per channel; low-magnitude bins gracefully fall back to the prior.

\paragraph{Velocity losses.}
For the two modalities,
\begin{align}
    \Lmag &= \E\!\left[\|v^{\text{mag}}_\theta - v^{*}_{\text{mag}}\|^2\right], \\
    \Lphasevel &= \E\!\left[w(|\Xt|)\,\|v^{\text{phase}}_\theta - v^{*}_{\text{phase}}\|^2\right].
\end{align}

\paragraph{Inter-channel phase consistency (IPD) loss.}
The losses above are all channel-wise, so inter-channel coherence would only emerge as a byproduct.
To make it an explicit training signal, we project predictions onto $\Sone$ by unit normalization, compute relative phase vectors $r_{ij} = e^{i(\phi_i - \phi_j)}$ for each channel pair $(i,j)$ from both the prediction and the target, and maximize their inner product:
\begin{equation}\label{eq:ipd_loss}
    \Lipd = \frac{1}{|\mathcal{P}|}\!\!\sum_{(i,j)\in\mathcal{P}} \E\!\left[w_{ij}\big(1 - \langle \hat r_{ij},\, r^{*}_{ij}\rangle\big)\right],
\end{equation}
where $\mathcal{P}$ is the set of channel pairs, $\langle\cdot,\cdot\rangle$ is the inner product of the two-dimensional unit vectors ($= \cos(\Delta\text{IPD})$), 
and $w_{ij} = \sigma\!\bigl((\min(\log m_i, \log m_j) - \mu)/s\bigr)$ applies the gating of (Eq.~\ref{eq:mag_weight}) with the element-wise minimum of the two channels' log-magnitudes in place of $\log|X^{(t)}|$, so a bin contributes only where both channels are energetic.
This loss is invariant to a common rotation of $\phi_i$ and $\phi_j$ (global phase shift invariance), matching the invariance of the physical quantities we evaluate (DOA and polarization direction) and thereby placing the learning signal directly on inter-channel structure.

\paragraph{Auxiliary losses.}
Velocity regression alone enforces neither the unit-circle constraint nor endpoint consistency.
From the one-step phase prediction $\hat z^{\text{phase}}_{1|t_p}=(\hat c,\hat s)$ we add $\Lphaseq$, a weighted sum of a phase direction loss, an instantaneous-frequency loss on $\partial_t\phi$, and a unit-norm penalty $(\hat c^2+\hat s^2-1)^2$; and $\Lrec = \E[\|\hat z^{\text{mag}}_{1|t_m} - \log|\Xt|\|^2]$, which reweights the magnitude velocity error by $(1-t_m)^2$ to emphasize early timesteps. Definitions in App.~\ref{app:implementaion_details}.

\paragraph{Total loss.}
The overall objective is
\begin{equation}\label{eq:total-loss}
    \mathcal{L} = \lambda_1 \Lmag + \lambda_2 \Lphasevel + \lambda_3 \Lphaseq + \lambda_4 \Lipd + \lambda_5 \Lrec.
\end{equation}
The weights $\lambda_\bullet$ follow a schedule across training stages (early stages prioritize velocity learning stability; later stages emphasize phase quality terms); specific values are given in the implementation details.

\begin{figure*}[t]
    \centering
    \begin{subfigure}[b]{0.49\textwidth}
        \centering
        \includegraphics[width=\textwidth]{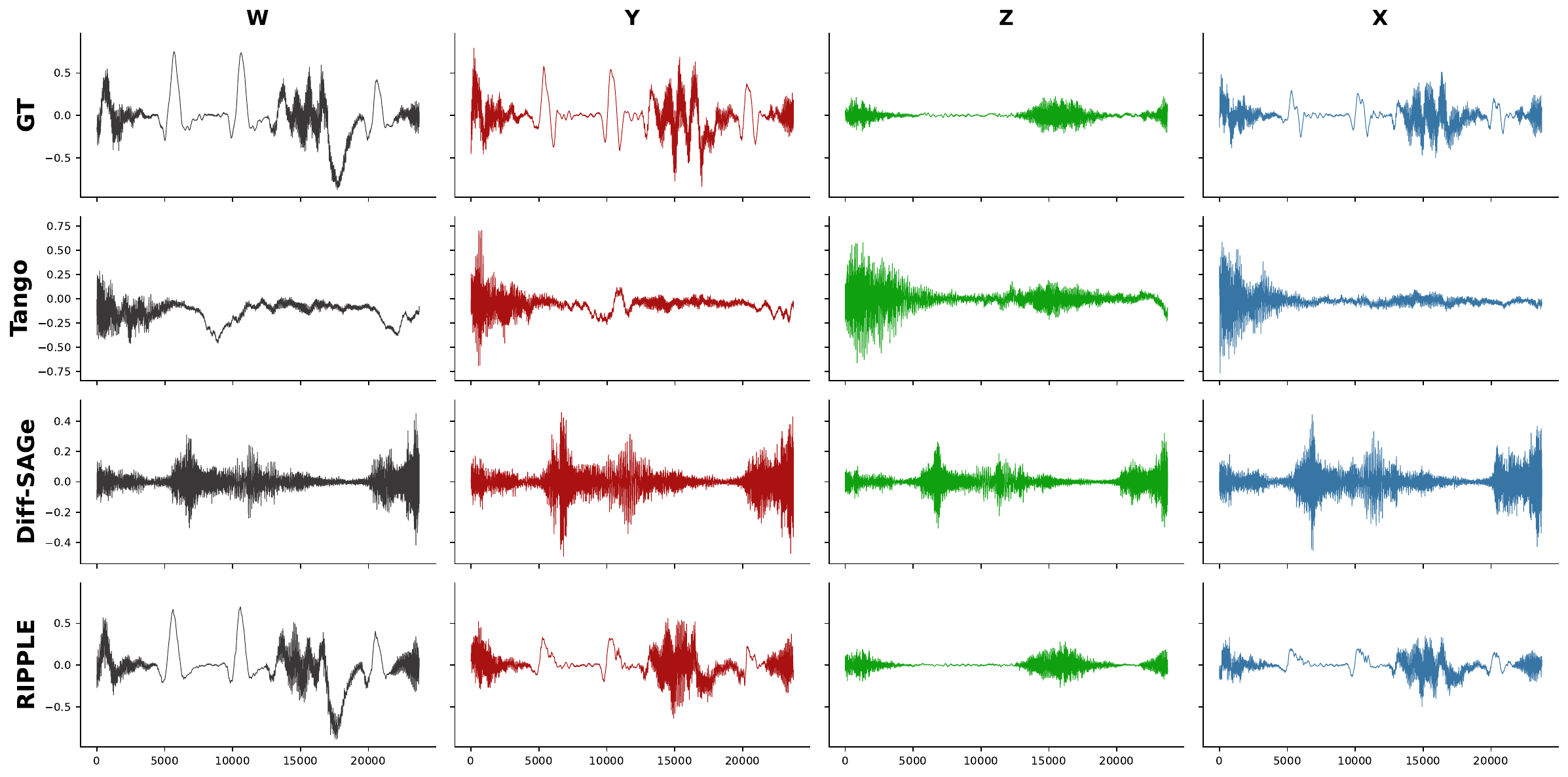}
        \caption{Direct baselines (no source-target adaptation).
        }
        \label{fig:waveform_direct}
    \end{subfigure}
    \hfill
    \begin{subfigure}[b]{0.49\textwidth}
        \centering
        \includegraphics[width=\textwidth]{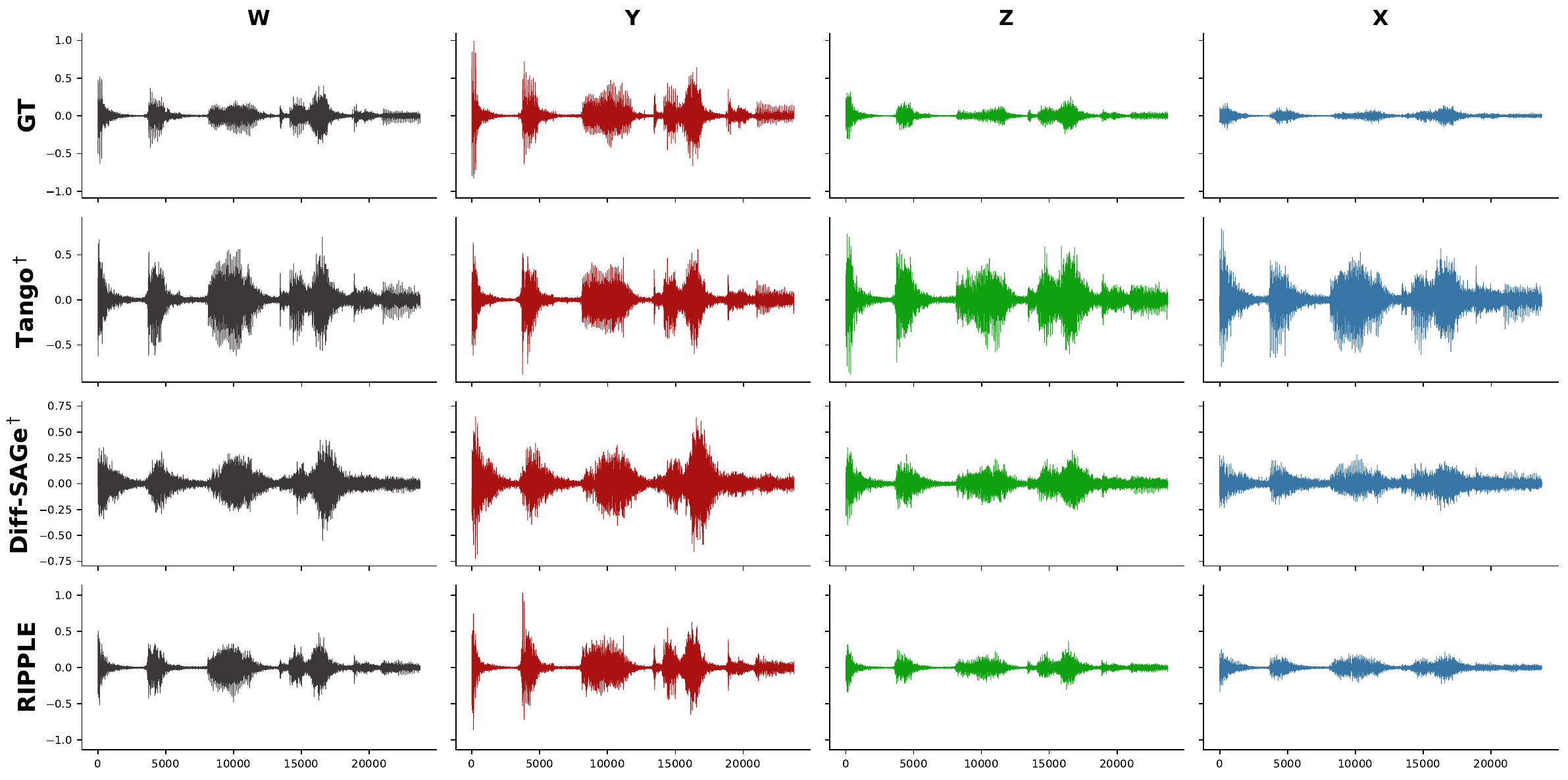}
        \caption{Source-target adapted baselines (marked $^\dagger$).}
        \label{fig:waveform_adapted}
    \end{subfigure}
    \caption{
    FOA waveforms (W/Y/Z/X) for a Spatial LibriSpeech test sample; rows share an amplitude scale. 
    }
    \label{fig:waveform_comparison}
\end{figure*}

\begin{table*}[t]
    \centering
    \caption{Spatial audio fidelity and spatial accuracy on Spatial LibriSpeech
    (cross-room); lower is better. Methods marked $^\dagger$ are baselines
    adapted to the source-to-target setting. ImmerseDiffusion codec
    reconstruction is analyzed in Appendix~\ref{app:base_adap}.}
    \label{tab:spatial_main}
    \setlength{\tabcolsep}{2.5pt}
    \small
    \begin{tabular}{llccccccc}
    \toprule
    \multirow{2}{*}{\textbf{Setting}}
    & \multirow{2}{*}{\textbf{Method}}
    & \multicolumn{4}{c}{\textbf{Fidelity}}
    & \multicolumn{3}{c}{\textbf{Spatial accuracy}} \\
    \cmidrule(lr){3-6} \cmidrule(lr){7-9}
    & & Wave L2 $\downarrow$ & Amp. L2 $\downarrow$ & Phase L2 $\downarrow$ & MRSTFT $\downarrow$
    & $L1_\theta$ $\downarrow$ & $L1_\varphi$ $\downarrow$ &  $\Delta_{\text{Spatial-Angle}}$ $\downarrow$ \\
    \midrule
    \multirow{1}{*}{GT}
    & GT reference (estimators on GT waveforms) & -- & -- & -- & -- & 0.139 & 0.127 & 0.208 \\
    \midrule
    \multirow{3}{*}{Direct}
    & Tango \cite{ghosal2023text}
        & 0.027 & 2.111 & 0.730 & 3.059 & 1.540 & 0.511 & 1.555 \\
    & ImmerseDiffusion \cite{heydari2025immersediffusion}
        & \textbf{0.013} & 1.007 & 0.994 & 2.905 & 1.523 & 0.592 & 1.507 \\
    & Diff-SAGe \cite{kushwaha2025diffsage}
        & 0.035 & 1.197 & 0.685 & 2.984 & 0.261 & 0.280 & 0.425 \\
    \midrule
    \multirow{2}{*}{Adapted}
    & Tango$^\dagger$ (mel + BigVGAN)
        & 0.034 & 1.057 & 1.545 & 1.654 & 1.542 & 0.461 & 1.557 \\
    & Diff-SAGe$^\dagger$
        & 0.020 & 0.550 & 1.363 & 2.214 & 0.455 & 0.218 & 0.534 \\
    \midrule
    \multirow{2}{*}{Ours}
    & RIPPLE (GL)
        & 0.022 & 0.460 & 1.072 & 1.318 & 1.591 & 0.395 & 1.586 \\
    & \textbf{RIPPLE}
        & 0.019 & \textbf{0.432} & \textbf{0.619} & \textbf{1.249}
        & \textbf{0.243} & \textbf{0.152} & \textbf{0.319} \\
    \bottomrule
    \end{tabular}
\end{table*}

\subsection{Training and Inference}
\label{subsec:training-and-infer}
Training regresses both velocities under the combined objective (Eq.~\eqref{eq:total-loss}).
At inference, the two ODEs are integrated \emph{sequentially} (Fig.~\ref{fig:main}): we first integrate the magnitude flow from $\log|\Xs|$ to a predicted target magnitude $\log|\hat\Xt|$, then construct the source-informed Griffin--Lim prior $p_0$ from the \emph{predicted} magnitude $\hat m$ and the source phase $\angle\Xs$ (Eq.~\ref{eq:p0}), and integrate the phase ODE from $p_0$.
Magnitude enters the phase stage at two points—the prior construction and the channel-wise conditioning of the phase head—and at inference the predicted $\hat{m}$ replaces the oracle $|\Xt|$ at both. This substitution is the only train/inference mismatch in the pipeline.
Pseudocode and integration details are in Appendix~\ref{app:algorithms}.

\section{Experimental Setup}

\paragraph{Datasets and pairs.}

\textbf{Primary: Spatial LibriSpeech} \citep{sarabia2023spatial}, 658.7 h of FOA speech; each utterance is rendered in one of 8,000+ synthetic rooms (4 ch, 16\,kHz) with source-position and room metadata. Cross-room pairs—two renderings of the same utterance—yield 79,232 training and ${\sim}$111K test directional pairs. \textbf{Secondary: SCEDC} \citep{scedc}, following the HEGGS protocol \citep{jung2025heggs} (60s Z/N/E waveforms at 100Hz, 4{,}016 train / 861 test events, cross-station pairs from the same event), with the bandpass widened to 0.1--45\,Hz to capture low-frequency ground motion. Details in Appendix \ref{app:pair_construction}.

\paragraph{Baselines.}
In the spatial-audio domain, we compare against two native spatial generators—ImmerseDiffusion \citep{heydari2025immersediffusion} and Diff-SAGe \citep{kushwaha2025diffsage}—and Tango \cite{ghosal2023text} with per-channel BigVGAN vocoding \citep{lee2022bigvgan}. Because these are not built for source-to-target translation, we also train source-adapted variants of Diff-SAGe and Tango (marked $^\dagger$), conditioned on the source spectrogram and trained on our pairs (Appendix~\ref{app:base_adap}); these isolate whether source conditioning alone transports inter-channel phase. To separate learned phase from classical recovery directly, RIPPLE (GL) keeps our magnitude head but replaces the phase head with standard random-initialized Griffin–Lim, as used in recovery-based pipelines \citep{jung2025heggs}; RIPPLE is the complete model. In the seismic domain, we compare against HEGGS \citep{jung2025heggs} and a GAN baseline \cite{wang2021seismogen}.

 \begin{table*}[t]
    \centering
    \caption{
    Cross-station translation on SCEDC (random-direction PAE: $57.3^\circ$; metric definitions in Appendix~\ref{app:metrics}.). 
    }
    \label{tab:seismic_main}
    \setlength{\tabcolsep}{2.5pt}
    \small
    \begin{tabular}{lccccccc|cccccc}
    \toprule
    \multirow{2}{*}{\textbf{Method}}
    & \multicolumn{7}{c|}{\textbf{Goodness-of-fit (GOF)} $\uparrow$}
    & \multicolumn{6}{c}{\textbf{Inter-component particle motion} $\downarrow$} \\
    \cmidrule(lr){2-8} \cmidrule(lr){9-14}
    & PGA & PSA & FAS & Env. & $I_A$ & $D_S$ & Mean
    & PAE & RLE & PAE$_{\mathrm{P}}$ & RLE$_{\mathrm{P}}$ & PAE$_{\mathrm{S}}$ & RLE$_{\mathrm{S}}$ \\
    \midrule
    GAN \cite{wang2021seismogen}
        & 19.17 & 22.14 & 22.78 & 26.70 & 9.77 & 19.46 & 20.00
        & 58.39 & 0.287 & 57.54 & 0.379 & 59.13 & 0.366 \\
    HEGGS \cite{jung2025heggs}
        & 36.41 & 40.06 & 36.84 & 85.74 & 31.48 & \textbf{68.56} & 49.85
        & 50.21 & 0.179 & 34.58 & 0.178 & 54.98 & 0.173 \\
    \midrule
    RIPPLE (GL)
        & 81.96 & 53.05 & \textbf{47.61} & 85.61 & 41.20 & 65.64 & 62.51
        & 49.79 & 0.169 & 40.42 & 0.181 & 51.87 & 0.153 \\
    \textbf{RIPPLE}
        & \textbf{90.56} & \textbf{55.86} & 47.01 & \textbf{85.75}
        & \textbf{67.73} & 65.85 & \textbf{68.79}
        & \textbf{32.57} & \textbf{0.111} & \textbf{26.48} & \textbf{0.121}
        & \textbf{33.81} & \textbf{0.098} \\
    \bottomrule
    \end{tabular}
\end{table*}

\paragraph{Metrics.}

Our central metrics measure inter-channel coherence directly; we report standard fidelity metrics alongside them to show the contrast our analysis predicts. In the spatial-audio domain, coherence is captured by spatial accuracy (following ImmerseDiffusion \cite{heydari2025immersediffusion}): azimuth $L1_\theta$, elevation $L1_\varphi$, and an overall $\Delta_{\text{Spatial-Angle}}$—which, since the intensity vector derives from the phase–amplitude alignment between W and Y/Z/X, directly measures the inter-channel coherence a downstream localizer consumes. Alongside these we report waveform/spectral fidelity (Wave L2, Amplitude L2, Phase L2, MRSTFT) following BinauralGrad \cite{leng2022binauralgrad}. In the seismic domain, the goodness-of-fit (GOF) protocol of \cite{olsen2010goodness}—per-component scores in [0, 100] over Z/N/E across six intensity measures—is amplitude-based, so our coherence metric is polarization analysis \citep{jurkevics1988polarization}: polarization angle error (PAE) and rectilinearity L1 error (RLE), reported also in windows at the TauP-predicted P/S arrivals (PAE$_{\mathrm{P}}$, PAE$_{\mathrm{S}}$). Definitions in Appendix~\ref{app:metrics}.

\section{Results}

\subsection{First-order Ambisonics Domain}
Table~\ref{tab:spatial_main} reports results on Spatial LibriSpeech; Figure~\ref{fig:waveform_direct} shows representative waveforms. \emph{RIPPLE} achieves the best scores on all phase-sensitive and spatial metrics. Notably, ImmerseDiffusion attains the lowest Wave L2 (0.013) while its azimuth error sits at the chance level ($1.523$ vs. $\pi/2 \approx 1.571$)—the starkest instance of the metric insensitivity our analysis predicts: a model can lead the table on waveform fidelity while carrying no directional information. This is not a training failure but a ceiling imposed upstream: its codec's reconstruction alone already lands azimuth at chance ($L1_\theta{=}1.526$; Appendix~\ref{app:base_adap}), and the full model coincides with that limit (1.523)—no generation in this latent can localize better than the codec permits. Against Diff-SAGe—the strongest direct baseline—the overall spatial error $\Delta_{\text{Spatial-Angle}}$ improves from 0.425 to 0.319, a 25\% relative reduction.

\paragraph{Phase head ablation: Generation beats recovery.}
Holding the magnitude path fixed, we replace the phase processing with each of two alternatives. \textbf{(a)} Griffin--Lim phase recovery raises Phase~L2 from $0.619$
to $1.072$ and $\Delta_{\text{Spatial-Angle}}$ from $0.319$ to $1.586$, a fivefold regression on spatial accuracy.
\textbf{(b)} A vocoder-based pipeline fares no better: Tango$^\dagger$, which renders its own predicted mel through BigVGAN, yields 1.545 and 1.557, respectively---worse than Griffin–Lim on phase fidelity despite a strong neural vocoder. This comparison varies the magnitude path as well as the phase path, so it is not a controlled phase-head ablation like (a).

\paragraph{Source-target adaptation alone does not transport phase.}
Adapting Diff-SAGe to the source-target setting (Diff-SAGe$^\dagger$) sharply improves its fidelity numbers (Wave~L2 $0.035 \to 0.020$, Amp.~L2 $1.197 \to 0.550$) but \emph{degrades} its Phase~L2 ($0.685 \to 1.363$): conditioning a complex-spectrogram model on the source spectrogram is not by itself sufficient to transport inter-channel phase coherence to the target (Fig.~\ref{fig:waveform_adapted}).
RIPPLE improves over Diff-SAGe$^\dagger$ on every metric (Table~\ref{tab:spatial_main}).

\subsection{Seismology Domain}

\begin{figure}[t]
    \centering
    \includegraphics[width=0.9\linewidth]{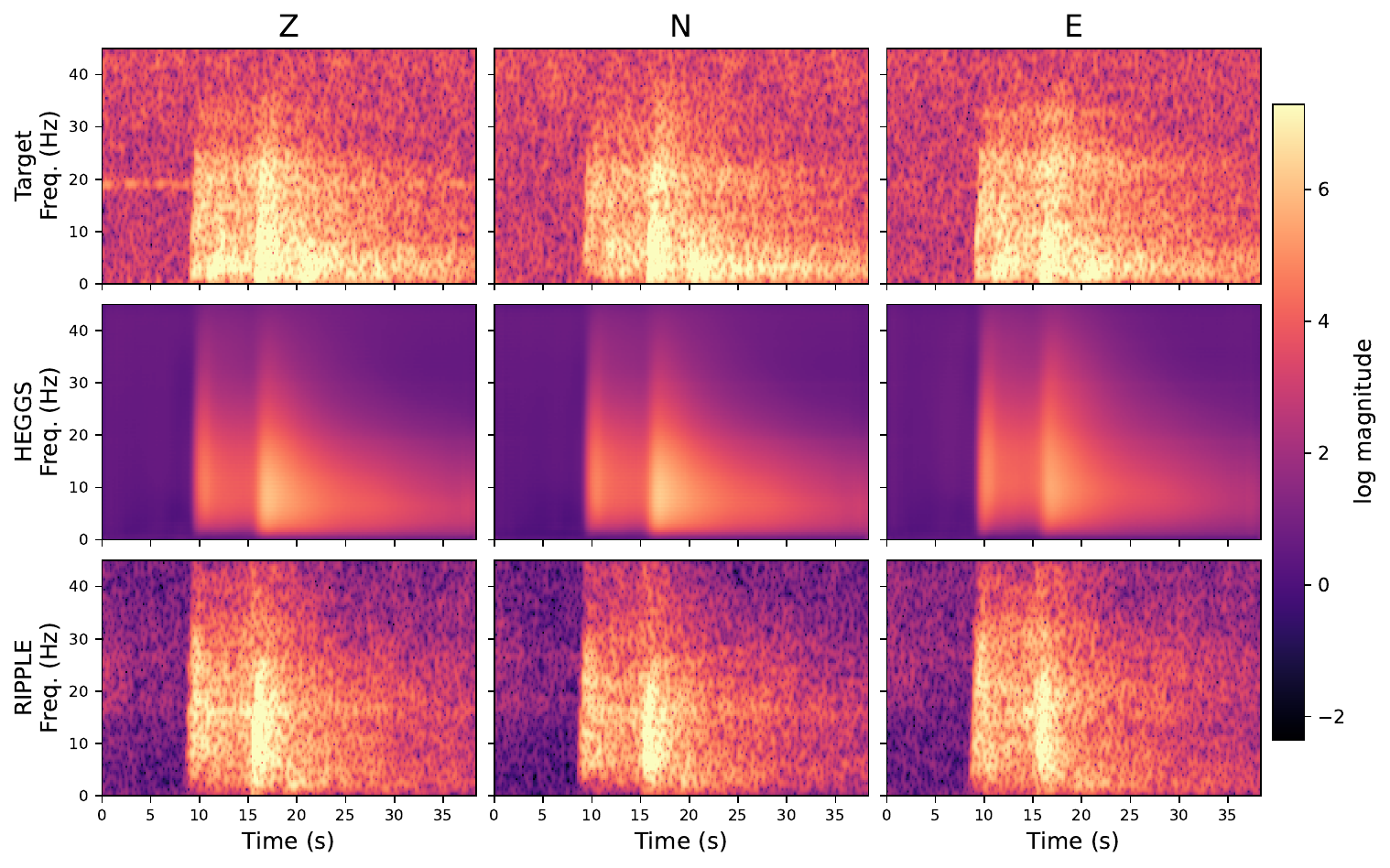}
    \caption{Three-component (Z/N/E) log-magnitude spectrograms for an SCEDC cross-station test sample.}
    \label{fig:seismic_spectrogram}
\end{figure}

Table~\ref{tab:seismic_main} reports cross-station translation on SCEDC, where all methods share the HEGGS conditioning protocol (Appendix~\ref{app:base_adap}) so the comparison isolates architecture. RIPPLE attains a mean GOF of 68.79 ("good" on the Olsen–Mayhew scale) against 49.85 ("fair") for HEGGS. The advantage concentrates on what HEGGS's compressed VAE latent cannot preserve---peak amplitude and energy content (PGA $36.41 \to 90.56$, $I_A$ $31.48 \to 67.73$), while the two are tied on envelope-level quantities (Env.\ 85.74 vs.\ 85.75); Figure~\ref{fig:seismic_spectrogram} and Appendix~\ref{app:add_visual} trace this to latent compression, and FAS (47.01) remains our lowest score.

\begin{table}[t]
    \centering
    \caption{Ablation on both domains, all else fixed (random-direction
    PAE: $57.3^\circ$; lower is better except GOF). Full metrics and row
    definitions in Appendix~\ref{app:ablations}.}
    \label{tab:ablation_merged}
    \setlength{\tabcolsep}{2pt}
    \small
    \begin{tabular}{l@{\hskip 3pt}ccc|ccc}
        \toprule
        & \multicolumn{3}{c|}{Spatial audio} & \multicolumn{3}{c}{Seismic} \\
        Configuration & Ph.\,L2 & $L1_\theta$ & $L1_\varphi$
                      & GOF$\uparrow$ & PAE & PAE$_\mathrm{S}$ \\
        \midrule
        \textbf{RIPPLE} & \textbf{0.619} & \textbf{0.243} & \textbf{0.152} & \textbf{68.79} & \textbf{32.57} & \textbf{33.81} \\
        w/o src-coherent init & 0.683 & 0.253 & 0.161 & 65.13 & 43.28 & 48.23 \\
        w/o GL prior & 1.404 & 1.616 & 0.615 & 59.39 & 52.07 & 57.33 \\
        w/o IPD ($\lambda_4{=}0$) & 1.451 & 1.627 & 0.188 & 60.04 & 52.38 & 57.31 \\
        coupled $t$ ($t_m{=}t_p$) & 0.807 & 1.582 & 0.200 & 57.84 & 50.98 & 50.28 \\
        \midrule
        RIPPLE (GL) & 1.072 & 1.591 & 0.395 & 62.51 & 49.79 & 51.87 \\
        prior only (no flow) & 0.717 & 1.459 & 0.291 & 62.87 & 45.17 & 44.58 \\
        \bottomrule
        \end{tabular}
\end{table}

\paragraph{Inter-component particle motion.}

The GOF above is amplitude-based; our central claim is tested only by polarization, and there the result is sharpest. S-wave polarization rides on the relative phase between horizontal components—exactly what per-channel recovery cannot produce—and every recovery-based method collapses to chance: GAN $59.1^\circ$, HEGGS $55.0^\circ$, RIPPLE (GL) $51.9^\circ$, against a $57.3^\circ$ random expectation. RIPPLE alone preserves it, at $33.8^\circ$. The contrast is mechanistic, not architectural: HEGGS and RIPPLE (GL) reach nearly identical polarization errors despite entirely different generators (LDM likewise; App.~\ref{app:base_adap}) because they share per-channel Griffin–Lim recovery—learning the phase is what breaks the ceiling ($49.8^\circ\!\to\!32.6^\circ$, RLE $0.169 \to 0.111$). The P/S split supports the mechanism: P-wave polarization is partly constrained by source–receiver geometry and thus recoverable from magnitudes (HEGGS $34.6^\circ$), while S-wave polarization is not—only learned phase recovers it (GOF-level corroboration in Appendix~\ref{app:ablations}).

\subsection{Ablation Studies}

Table~\ref{tab:ablation_merged} removes one component at a time from \emph{RIPPLE} on both domains, keeping the magnitude path and all other settings fixed (full metrics in Appendix~\ref{app:ablations}).

\begin{figure}[t]
    \centering
    \includegraphics[width=0.6\linewidth]{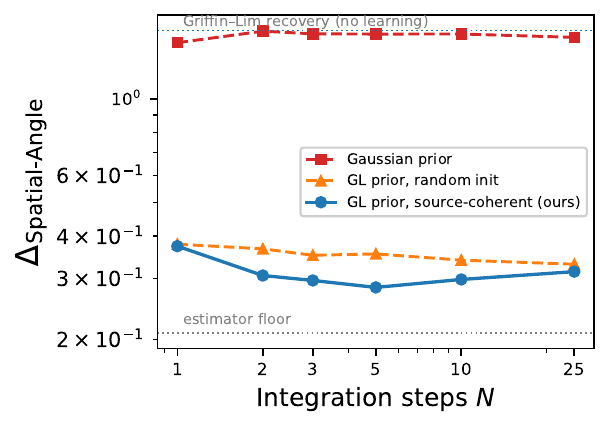}
    \caption{$\Delta_{\text{Spatial-Angle}}$  vs. integration steps $N$ on Spatial LibriSpeech (100 test pairs; 4N network evaluations total, 2N per Heun stage). Dashed line: classical Griffin–Lim recovery.}
    \label{fig:nfe_curve}
\end{figure}

\paragraph{A structured prior determines feasibility.}
Replacing the Griffin--Lim prior with a Gaussian collapses both domains to their no-learning baselines: in the spatial-audio domain, $\Delta_{\text{Spatial-Angle}}$ regresses from $0.319$ to $1.604$ (the level of classical Griffin--Lim recovery), and in the seismic domain, PAE$_\mathrm{S}$ lands at the random-direction expectation ($57.33^\circ$ vs.\ $57.3^\circ$). Without a structured starting point, and this holds at every integration budget (Fig.~\ref{fig:nfe_curve}).

\paragraph{Source coherence matters where magnitudes cannot reach.}

The endpoint contribution of source-coherent initialization differs across domains, validating our analysis. In the spatial-audio domain, removing it barely moves the endpoint ($0.319 \to 0.332$); its role is few-step efficiency (at $N{=}2$ it already beats the randomly initialized prior at $N{=}25$; Figure~\ref{fig:nfe_curve}). In the seismic domain, the endpoint regresses (PAE $32.6^\circ \to 43.3^\circ$), and the degradation is concentrated exactly where phase is unrecoverable from magnitudes: P-wave polarization degrades by only $6.7^\circ$ ($26.5^\circ \to 33.2^\circ$), while S-wave polarization degrades by more than twice as much ($33.8^\circ \to 48.2^\circ$). Even without the flow, the prior alone already sits well below chance on S-wave polarization (PAE$_\mathrm{S}$ $44.6^\circ$ vs.\ $57.3^\circ$; Table~\ref{tab:ablation_merged}, \emph{prior only})---initialization preserves exactly what the transfer functions preserve, and no more.

\paragraph{Cross-channel coherence requires both the IPD objective and decoupled timesteps.}
Setting $\lambda_4{=}0$ leaves elevation nearly intact ($0.152 \to 0.188$) but collapses cross-channel structure: azimuth collapses ($L1_\theta$ $0.243 \to 1.627$), and in the seismic domain, PAE$_\mathrm{S}$ again reaches the random level ($57.31^\circ$)---channel-wise velocity losses alone do not yield coherence as a byproduct. Coupling the timesteps ($t_m{=}t_p$) exposes the same fault line through a different mechanism: per-bin phase is largely spared (Phase L2 $0.619 \to 0.807$) while inter-channel coherence collapses ($\Delta_{\text{Spatial-Angle}}$ $0.319 \to 1.581$), consistent with the noise-level correlation induced by the shared encoder when $t_m{=}t_p$.

\begin{table}[t]
    \centering
    \caption{Oracle reconstruction of ground-truth targets---no
    generation, no magnitude error. Fidelity survives every recovery
    route; the coherence metric does not. GT estimator floor:
    $L1_\theta=0.139$. Full results in Appendix~\ref{app:base_adap}.}
    \label{tab:oracle}
    \setlength{\tabcolsep}{2pt}
    \small
    \begin{tabular}{lcc}
    \toprule
    Recovery route & Coherence & Fidelity \\
    \midrule
    \textit{FOA} & $L1_\theta\downarrow$ [chance 1.571] & Wave L2 $\downarrow$ \\
    \midrule
    Griffin--Lim, GT magnitudes & 1.567 & 0.013 \\
    BigVGAN round trip          & 1.607 & 0.035 \\
    ImmerseDiffusion codec      & 1.526 & 0.011 \\
    \midrule
    \textit{Seismic} & PAE$_\mathrm{S}\downarrow$ [chance 57.3$^\circ$] & GOF $\uparrow$ \\
    \midrule
    VAE round trip $+$ GT phase & $3.7^\circ$  & 98.4 \\
    VAE round trip $+$ GL phase & $42.9^\circ$ & 95.3 \\
    \bottomrule
    \end{tabular}
\end{table}

\section{Discussion}

One might attribute the gains to Griffin--Lim merely supplying a better magnitude-conditioned starting point rather than to inter-channel structure. The ablation ladder separates the two: randomizing the prior's initialization while keeping Griffin--Lim isolates the source-coherent contribution, and the further collapse under a Gaussian prior isolates magnitude consistency (Table~\ref{tab:ablation_merged}, Figure~\ref{fig:nfe_curve}). 
A single distinction remains: the Gaussian replacement removes magnitude consistency and valid $\mathbb{S}^1$ structure together. The $K=0$ prior of Appendix~\ref{app:k_sensitivity}---the source phase with no magnitude projection---partially separates them and lands far closer to the full model than to the Gaussian level; a flow trained with such a prior would settle it.

\paragraph{The bottleneck is the recovery stage, not the generator.} 
Table~\ref{tab:oracle} removes the generator from the question: every row reconstructs ground-truth targets, so any failure is due to recovery alone. Fidelity survives every route while the quantity that downstream analyses consume collapses: azimuth reaches chance on all three spatial routes, and replacing ground-truth phase with recovery degrades S-wave polarization by an order of magnitude. The ceilings bind tightly—a latent-codec generator's spatial accuracy coincides with its own codec's limit (1.523 vs.\ 1.526), and three architecturally distinct seismic generators sharing per-channel recovery land within a few degrees ($55.0^\circ, 51.9^\circ, 56.8^\circ$). Nor is it a magnitude problem: recovery from near-oracle magnitudes ($42.9^\circ$) stays worse than RIPPLE on its own predicted ones ($33.8^\circ$). A recovery-based pipeline inherits the ceiling of its recovery stage; a better generator cannot lift it, but learned phase can.

\paragraph{Limitations.}
First, the framework presumes that source and target share inter-channel structure through a common source signal. When this
assumption weakens---as in low-SNR recordings, or in pairs whose transfer functions destroy the shared structure---the prior
degenerates toward a randomly initialized one, and the \emph{w/o source-coherent init} ablation (Table~\ref{tab:ablation_merged}) measures the cost: efficiency loss where the target's structure is recoverable from magnitudes, and endpoint loss where it is not (Figure~\ref{fig:nfe_curve}). Second, our evaluation is objective (DOA, polarization): FOA is not a playback format, so a listening test would entangle inter-channel structure with the binaural
renderer and HRTF \citep{wenzel1993localization}. Finally, $K$ trades prior quality against compute; we train at a fixed $K$ and
show inference-time robustness only (Appendix~\ref{app:k_sensitivity}).
Source-free tasks such as text-to-spatial generation remain future work.

\section{Conclusion}
Multi-channel waveforms carry their physical content---source direction in ambisonics, particle motion in seismograms---in the relationships between channels, yet existing generative AI methods have delegated phase to channel-independent recovery. We show that this delegation carries a measurable cost, and offer a practical remedy: reinterpreting Griffin--Lim as a source-coherent prior for a decoupled rectified flow turns phase into a generation target without discarding what classical estimation provides. This approach improves empirical performance on first-order ambisonics environment transfers and seismic cross-station translations---most sharply on S-wave polarization, which per-channel reconstruction leaves near chance level. 
We hope this reframing---\emph{generating multi-channel phase, not recovering it}---encourages treating inter-channel coherence as a first-class generation target, tracked alongside magnitude-based fidelity.

\bibliography{references_try1}

@inproceedings{kushwaha2025diffsage,
  title     = {{Diff-SAGe}: End-to-End Spatial Audio Generation Using Diffusion Models},
  author    = {Kushwaha, Saksham Singh and Ma, Jianbo and Thomas, Mark R. P. and Tian, Yapeng and Bruni, Avery},
  booktitle = {Proc. IEEE International Conference on Acoustics, Speech and Signal Processing (ICASSP)},
  year      = {2025}
}

@inproceedings{heydari2025immersediffusion,
  title     = {{ImmerseDiffusion}: A Generative Spatial Audio Latent Diffusion Model},
  author    = {Heydari, Mojtaba and Souden, Mehrez and Conejo, Bruno and Atkins, Joshua},
  booktitle = {Proc. IEEE International Conference on Acoustics, Speech and Signal Processing (ICASSP)},
  year      = {2025}
}

@inproceedings{leng2022binauralgrad,
  title     = {{BinauralGrad}: A Two-Stage Conditional Diffusion Probabilistic Model for Binaural Audio Synthesis},
  author    = {Leng, Yichong and Chen, Zehua and Guo, Junliang and Liu, Haohe and Chen, Jiawei and Tan, Xu and Mandic, Danilo P. and He, Lei and Li, Xiang-Yang and Qin, Tao and Zhao, Sheng and Liu, Tie-Yan},
  booktitle = {Advances in Neural Information Processing Systems (NeurIPS)},
  year      = {2022}
}

@inproceedings{richard2021warpnet,
  title     = {Neural Synthesis of Binaural Speech from Mono Audio},
  author    = {Richard, Alexander and Markovic, Dejan and Gebru, Israel D. and Krenn, Steven and Butler, Gladstone Alexander and De la Torre, Fernando and Sheikh, Yaser},
  booktitle = {International Conference on Learning Representations (ICLR)},
  year      = {2021}
}

@inproceedings{lee2023nfs,
  title     = {Neural {Fourier} Shift for Binaural Speech Rendering},
  author    = {Lee, Jin Woo and Lee, Kyogu},
  booktitle = {Proc. IEEE International Conference on Acoustics, Speech and Signal Processing (ICASSP)},
  year      = {2023}
}

@inproceedings{sarabia2023spatial,
  title     = {Spatial {LibriSpeech}: An Augmented Dataset for Spatial Audio Learning},
  author    = {Sarabia, Miguel and Menyaylenko, Elena and Toso, Alessandro and Seto, Skyler and Aldeneh, Zakaria and Pirhosseinloo, Shadi and Zappella, Luca and Theobald, Barry-John and Apostoloff, Nicholas and Sheaffer, Jonathan},
  booktitle = {Proc. Interspeech},
  pages     = {3724--3728},
  year      = {2023},
  doi       = {10.21437/Interspeech.2023-2117}
}

@inproceedings{jung2025heggs,
  title     = {Broadband Ground Motion Synthesis by Diffusion Model with Minimal Condition},
  author    = {Jung, Jaeheun and Lee, Jaehyuk and Jung, Chang-Hae and Kim, Hanyoung and Jung, Bosung and Lee, Donghun},
  booktitle = {Proc. International Conference on Machine Learning (ICML)},
  year      = {2025}
}

@article{bi2025advancing,
  title={Advancing data-driven broadband seismic wavefield simulation with multiconditional diffusion model},
  author={Bi, Zhengfa and Nakata, Nori and Nakata, Rie and Ren, Pu and Wu, Xinming and Mahoney, Michael W},
  journal={IEEE Transactions on Geoscience and Remote Sensing},
  volume={63},
  pages={1--9},
  year={2025},
  publisher={IEEE}
}

@article{wang2021seismogen,
  title   = {{SeismoGen}: Seismic Waveform Synthesis Using {GAN} with Application to Seismic Data Augmentation},
  author  = {Wang, Tiantong and Trugman, Daniel T. and Lin, Youzuo},
  journal = {Journal of Geophysical Research: Solid Earth},
  volume  = {126},
  number  = {4},
  pages   = {e2020JB020077},
  year    = {2021},
  doi     = {10.1029/2020JB020077}
}

@article{li2024conseisgen,
  title   = {{ConSeisGen}: Controllable Synthetic Seismic Waveform Generation},
  author  = {Li, Yuanming and Yoon, Dongsik and Ku, Bonhwa and Ko, Hanseok},
  journal = {IEEE Geoscience and Remote Sensing Letters},
  volume  = {21},
  pages   = {1--5},
  year    = {2024},
  doi     = {10.1109/LGRS.2023.3338652}
}

@misc{scedc,
  title        = {Southern {California} Earthquake Data Center},
  author       = {{SCEDC}},
  howpublished = {Caltech. Dataset},
  year         = {2013},
  doi          = {10.7909/C3WD3xH1}
}

@article{olsen2010goodness,
  title={Goodness-of-fit criteria for broadband synthetic seismograms, with application to the 2008 Mw 5.4 Chino Hills, California, earthquake},
  author={Olsen, Kim B and Mayhew, John E},
  journal={Seismological Research Letters},
  volume={81},
  number={5},
  pages={715--723},
  year={2010},
  publisher={Seismological Society of America}
}

@article{jurkevics1988polarization,
  title={Polarization analysis of three-component array data},
  author={Jurkevics, Andy},
  journal={Bulletin of the seismological society of America},
  volume={78},
  number={5},
  pages={1725--1743},
  year={1988},
  publisher={The Seismological Society of America}
}

@inproceedings{liu2023rectified,
  title     = {Flow Straight and Fast: Learning to Generate and Transfer Data with Rectified Flow},
  author    = {Liu, Xingchao and Gong, Chengyue and Liu, Qiang},
  booktitle = {International Conference on Learning Representations (ICLR)},
  year      = {2023}
}

@inproceedings{lipman2023flow,
  title     = {Flow Matching for Generative Modeling},
  author    = {Lipman, Yaron and Chen, Ricky T. Q. and Ben-Hamu, Heli and Nickel, Maximilian and Le, Matt},
  booktitle = {International Conference on Learning Representations (ICLR)},
  year      = {2023}
}

@inproceedings{albergo2023interpolants,
  title     = {Building Normalizing Flows with Stochastic Interpolants},
  author    = {Albergo, Michael S. and Vanden-Eijnden, Eric},
  booktitle = {International Conference on Learning Representations (ICLR)},
  year      = {2023}
}

@inproceedings{rombach2022ldm,
  title     = {High-Resolution Image Synthesis with Latent Diffusion Models},
  author    = {Rombach, Robin and Blattmann, Andreas and Lorenz, Dominik and Esser, Patrick and Ommer, Bj{\"o}rn},
  booktitle = {Proc. IEEE/CVF Conference on Computer Vision and Pattern Recognition (CVPR)},
  pages     = {10684--10695},
  year      = {2022}
}

@article{griffin1984,
  title   = {Signal Estimation from Modified Short-Time {Fourier} Transform},
  author  = {Griffin, Daniel W. and Lim, Jae S.},
  journal = {IEEE Transactions on Acoustics, Speech, and Signal Processing},
  volume  = {32},
  number  = {2},
  pages   = {236--243},
  year    = {1984},
  doi     = {10.1109/TASSP.1984.1164317}
}

@inproceedings{kong2020hifigan,
  title     = {{HiFi-GAN}: Generative Adversarial Networks for Efficient and High Fidelity Speech Synthesis},
  author    = {Kong, Jungil and Kim, Jaehyeon and Bae, Jaekyoung},
  booktitle = {Advances in Neural Information Processing Systems (NeurIPS)},
  year      = {2020}
}

@inproceedings{liu2025omniaudio,
  title     = {{OmniAudio}: Generating Spatial Audio from 360-Degree Video},
  author    = {Liu, Huadai and Luo, Tianyi and Jiang, Qikai and Luo, Kaicheng and Sun, Peiwen and Wan, Jialei and Huang, Rongjie and Chen, Qian and Wang, Wen and Li, Xiangtai and Zhang, Shiliang and Yan, Zhijie and Zhao, Zhou and Xue, Wei},
  booktitle = {Proc. International Conference on Machine Learning (ICML)},
  year      = {2025}
}

@article{templin2025sonicmotion,
  title   = {{SonicMotion}: Dynamic Spatial Audio Soundscapes with Latent Diffusion Models},
  author  = {Templin, Christian and Zhu, Yanda and Wang, Hao},
  journal = {arXiv preprint arXiv:2507.07318},
  year    = {2025}
}

@article{sudarsanam2025foatokenizer,
  title={FOA Tokenizer: Low-bitrate Neural Codec for First Order Ambisonics with Spatial Consistency Loss},
  author={Sudarsanam, Parthasaarathy and Braun, Sebastian and Gamper, Hannes},
  journal={arXiv preprint arXiv:2510.22241},
  year={2025}
}

@inproceedings{siuzdak2024vocos,
  title     = {{Vocos}: Closing the Gap between Time-Domain and {Fourier}-Based Neural Vocoders for High-Quality Audio Synthesis},
  author    = {Siuzdak, Hubert},
  booktitle = {International Conference on Learning Representations (ICLR)},
  year      = {2024}
}

@inproceedings{ghosal2023text,
  title={Text-to-audio generation using instruction guided latent diffusion model},
  author={Ghosal, Deepanway and Majumder, Navonil and Mehrish, Ambuj and Poria, Soujanya},
  booktitle={Proceedings of the 31st ACM international conference on multimedia},
  pages={3590--3598},
  year={2023}
}

@inproceedings{lee2022bigvgan,
  title={{BigVGAN}: A Universal Neural Vocoder with Large-Scale Training},
  author={Lee, Sang Gil and Ping, Wei and Ginsburg, Boris and Catanzaro, Bryan and Yoon, Sungroh},
  booktitle={11th International Conference on Learning Representations, ICLR 2023},
  year={2023}
}

@inproceedings{liang2025binauralflow,
  title={BinauralFlow: A Causal and Streamable Approach for High-Quality Binaural Speech Synthesis with Flow Matching Models},
  author={Liang, Susan and Markovic, Dejan and Gebru, Israel D and Krenn, Steven and Keebler, Todd and Sandakly, Jacob and Yu, Frank and Hassel, Samuel and Xu, Chenliang and Richard, Alexander},
  booktitle={International Conference on Machine Learning},
  pages={37281--37298},
  year={2025},
  organization={PMLR}
}

@article{mathieu2020riemannian,
  title={Riemannian continuous normalizing flows},
  author={Mathieu, Emile and Nickel, Maximilian},
  journal={Advances in neural information processing systems},
  volume={33},
  pages={2503--2515},
  year={2020}
}

@inproceedings{chen2024flow,
  title={Flow matching on general geometries},
  author={Chen, Ricky TQ and Lipman, Yaron},
  booktitle={International Conference on Learning Representations},
  volume={2024},
  pages={47922--47945},
  year={2024}
}

@inproceedings{guan2024lafma,
  title     = {{LAFMA}: A Latent Flow Matching Model for Text-to-Audio Generation},
  author    = {Guan, Wenhao and Wang, Kaidi and Zhou, Wangjin and Wang, Yang and Deng, Feng and Wang, Hui and Li, Lin and Hong, Qingyang and Qin, Yong},
  booktitle = {Proc. Interspeech},
  pages     = {4813--4817},
  year      = {2024},
  doi       = {10.21437/Interspeech.2024-1848}
}

@article{bergmeister2025highfem,
  title   = {High Resolution Seismic Waveform Generation Using Denoising Diffusion},
  author  = {Bergmeister, Andreas and Palgunadi, Kadek Hendrawan and Bosisio, Andrea and Ermert, Laura and Koroni, Maria and Perraudin, Nathana{\"e}l and Dirmeier, Simon and Meier, Men-Andrin},
  journal = {Journal of Geophysical Research: Machine Learning and Computation},
  year    = {2025}
}

@inproceedings{liu2023i2sb,
  title     = {{I$^2$SB}: Image-to-Image {S}chr{\"o}dinger Bridge},
  author    = {Liu, Guan-Horng and Vahdat, Arash and Huang, De-An and
               Theodorou, Evangelos A. and Nie, Weili and Anandkumar, Anima},
  booktitle = {Proceedings of the 40th International Conference on Machine Learning (ICML)},
  year      = {2023}
}

@inproceedings{ICLR2024_20e45668,
 author = {Zhou, Linqi and Lou, Aaron and Khanna, Samar and Ermon, Stefano},
 booktitle = {International Conference on Learning Representations},
 editor = {B. Kim and Y. Yue and S. Chaudhuri and K. Fragkiadaki and M. Khan and Y. Sun},
 pages = {8160--8171},
 title = {Denoising Diffusion Bridge Models},
 url = {https://proceedings.iclr.cc/paper_files/paper/2024/file/20e45668fefa793bd9f2edf19be12c4b-Paper-Conference.pdf},
 volume = {2024},
 year = {2024}
}

@inproceedings{bao2023one,
  title={One transformer fits all distributions in multi-modal diffusion at scale},
  author={Bao, Fan and Nie, Shen and Xue, Kaiwen and Li, Chongxuan and Pu, Shi and Wang, Yaole and Yue, Gang and Cao, Yue and Su, Hang and Zhu, Jun},
  booktitle={International Conference on Machine Learning},
  pages={1692--1717},
  year={2023},
  organization={PMLR}
}

@article{wenzel1993localization,
  title={Localization using nonindividualized head-related transfer functions},
  author={Wenzel, Elizabeth M and Arruda, Marianne and Kistler, Doris J and Wightman, Frederic L},
  journal={The Journal of the Acoustical Society of America},
  volume={94},
  number={1},
  pages={111--123},
  year={1993},
  publisher={Acoustical Society of America}
}

@article{kumar2023high,
  title={High-fidelity audio compression with improved rvqgan},
  author={Kumar, Rithesh and Seetharaman, Prem and Luebs, Alejandro and Kumar, Ishaan and Kumar, Kundan},
  journal={Advances in Neural Information Processing Systems},
  volume={36},
  pages={27980--27993},
  year={2023}
}
\onecolumn
\newpage

\appendix
\setcounter{secnumdepth}{2}
\newpage

\section{Method Positioning}\label{app:positioning}
Table \ref{tab:novelty} shows the scoped characterization of the methods surveyed in related works. It is intended to make the comparison axes explicit; a dash indicates that the corresponding property is not established by the cited source.

\begin{table*}[t]
\centering
\caption{Positioning of the closest surveyed work. \textbf{Phase}: \emph{Rec}
delegates phase to a recovery module; \emph{impl.} generates waveforms without
parameterizing phase separately; \emph{Gen} treats phase as an explicit
generation target. \textbf{Src.\ coh.}: the input carries inter-channel
coherence to transport. \textbf{X-loss}: a cross-channel objective is trained.
\textbf{Coh.\ eval}: coherence-sensitive evaluation is reported.
``--'' denotes a property not established by the cited source. Several methods
generate phase; within this comparison RIPPLE is the only one that pairs it
with source coherence and an explicit cross-channel objective. References are
keyed by superscript below the table.}
\label{tab:novelty}
\footnotesize
\setlength{\tabcolsep}{5pt}
\renewcommand{\arraystretch}{1.1}
\begin{tabular}{@{}llclcccc@{}}
\toprule
\textbf{Method} & \textbf{Input\,$\to$\,output} & \textbf{Ch.} & \textbf{Repr.}
& \textbf{Phase} & \textbf{Src.\ coh.} & \textbf{X-loss} & \textbf{Coh.\ eval} \\
\midrule
\multicolumn{8}{@{}l}{\emph{Spatial audio}} \\
Mono$\to$binaural\textsuperscript{1}      & mono $\to$ 2-ch          & 2    & waveform          & impl. & $\times$ & --          & -- \\
BinauralFlow\textsuperscript{2}           & mono $\to$ 2-ch          & 2    & complex STFT      & Gen   & $\times$ & $\times$    & $\triangle$ \\
Video/lang.-cond.\textsuperscript{3}      & video/text $\to$ spatial & 2--4 & varies            & --    & $\times$ & $\times$    & -- \\
Mono-synth + vocoder\textsuperscript{4}   & text $\to$ FOA           & 4    & mel + voc.        & Rec   & $\times$ & $\times$    & $\times$ \\
ImmerseDiffusion\textsuperscript{5}       & text $\to$ FOA           & 4    & codec latent      & Rec   & $\times$ & $\times$    & $\times$ \\
Diff-SAGe\textsuperscript{6}              & meta./label $\to$ FOA       & 4    & complex STFT      & Gen   & $\times$ & $\times$    & $\triangle$ \\
FOA Tokenizer\textsuperscript{7}          & FOA $\to$ FOA            & 4    & neural codec      & Rec   & n/a      & $\triangle$ & -- \\
\midrule
\multicolumn{8}{@{}l}{\emph{Seismic}} \\
Conditional GANs\textsuperscript{8}       & label $\to$ 3-c          & 3    & waveform          & impl. & $\times$ & $\times$    & -- \\
Latent diffusion\textsuperscript{9}       & meta./src $\to$ 3-c      & 3    & amp.\ spec.       & Rec   & --       & $\times$    & -- \\
HEGGS\textsuperscript{10}                 & 3-c $\to$ 3-c            & 3    & amp.\ spec.\ + GL & Rec   & \checkmark & $\times$  & $\times$ \\
\midrule
\textbf{RIPPLE} (ours) & 4-c/3-c $\to$ same & 4/3 & log-mag $+\,\mathbb{S}^1$ phase
& \textbf{Gen} & \checkmark & \checkmark & \checkmark \\
\bottomrule
\end{tabular}

\vspace{2pt}
{\scriptsize\raggedright
\textsuperscript{1}\citet{richard2021warpnet,leng2022binauralgrad,lee2023nfs};
\textsuperscript{2}\citet{liang2025binauralflow};
\textsuperscript{3}\citet{liu2025omniaudio,templin2025sonicmotion};
\textsuperscript{4}\citet{ghosal2023text,lee2022bigvgan};
\textsuperscript{5}\citet{heydari2025immersediffusion};
\textsuperscript{6}\citet{kushwaha2025diffsage};
\textsuperscript{7}\citet{sudarsanam2025foatokenizer};
\textsuperscript{8}\citet{wang2021seismogen,li2024conseisgen};
\textsuperscript{9}\citet{bergmeister2025highfem,bi2025advancing};
\textsuperscript{10}\citet{jung2025heggs}.\par}
\end{table*}

\section{Experimental Details}\label{app:exp_details}
\subsection{Pair construction.}\label{app:pair_construction}
\textbf{Spatial LibriSpeech (cross-room).} 
Each utterance is rendered multiple times, each rendering placed in a different room; we form cross-room pairs by taking two renderings of the same utterance as $(\src,\tgt)$.
Train: 39{,}616 utterances rendered exactly twice $\rightarrow$ 79{,}232 directional pairs.
Test: 5{,}559 utterances rendered $\geq 7$ times (mostly 9) $\rightarrow$ $\sim$111K directional pairs, capped at 20 per utterance.
Train and test utterance sets are disjoint by construction (an utterance rendered exactly twice cannot appear in the $\geq 7$ rendering test pool); the 20-pair cap is a uniform random subsample per utterance with a fixed seed.
This matches our framing of one source through two transfer functions.
\textbf{SCEDC (cross-station, HEGGS-compatible).} 
As in HEGGS, we sample two stations of the same event as $(\src,\tgt)$ and define dataset size by the number of events, so per-epoch iterations scale with events rather than all pairs.
Following the HEGGS pairing scheme (two stations of one event as source/target), we rebuild the split from the SCEDC catalog, yielding 4,016 train / 861 test events.
 
\subsection{Baseline adaptation.}\label{app:base_adap}

\subsubsection{Spatial Baselines.}
Diff-SAGe and Tango are not designed for source-to-target translation; we extend each by additionally conditioning the generator on the source spectrogram, trained on the same pair construction as our model (variants marked $^\dagger$).

\paragraph{ImmerseDiffusion codec and full model.}
We reproduced the ImmerseDiffusion FOA codec---the DAC autoencoder \citep{kumar2023high} with its RVQ bottleneck replaced by a continuous VAE, at 128x compression, following \cite{heydari2025immersediffusion}---and evaluated both its reconstruction (autoencoding round trip, no generation) and the full generative model trained on our cross-room protocol, on the same Spatial LibriSpeech test targets as Table~\ref{tab:spatial_main} (Table~\ref{tab:immerse_codec}). Codec reconstruction alone lands azimuth near the chance level: $L1_\theta = 1.526$, where $\pi/2 \approx 1.571$ is the expected circular $L1$ error of any azimuth estimate independent of the target (estimator floor on GT waveforms: $0.139$); the original paper's published variants report $1.43$--$1.57$ on Spatial AudioCaps, consistent with this level. Notably, ImmerseDiffusion trains its FOA codec with per-channel MR-STFT losses, explicitly arguing that channel orthogonality removes the need for a cross-channel objective, and its codec reconstruction alone lands azimuth at chance ($L1_\theta=1.526$), a direct consequence of that channel-independent design.
Elevation, partially recoverable from channel energy ratios, degrades but survives the round trip ($L1_\varphi = 0.590$ vs.\ floor $0.127$)---the codec-level counterpart of the magnitude-recoverable/phase-dependent split between P- and S-wave
polarization (Table~\ref{tab:seismic_main}).

The full model then confirms this bound is tight: its spatial accuracy ($L1_\theta = 1.523$, $\Delta_{\text{SA}} = 1.508$) coincides with its own codec's reconstruction limit ($1.526$, $1.512$). The generator is not failing to learn; the codec caps what any generation in its latent space can localize. For the same reason we do not train a source-adapted ImmerseDiffusion$^\dagger$: source conditioning enters before the codec
and cannot lift this ceiling.

\begin{table}[H]
    \centering
    \caption{ImmerseDiffusion FOA codec \emph{reconstruction} (autoencoding round trip, no generation): published results vs.\ our reproduction.
    Upper block quoted from \citet{heydari2025immersediffusion} (evaluated on Spatial AudioCaps; trained on their full data mixture);
    lower block is our reproduction of the 128$\times$ codec, trained and evaluated on Spatial LibriSpeech only (Table~\ref{tab:spatial_main}).
    Estimator floors are dataset-specific, so absolute values are comparable within a block; the azimuth chance level is not---$\pi/2$ is the expected circular $L1$ error of any azimuth estimate independent of the target.
    All published variants sit at or near this level, as does our reproduction. Lower is better.}
    \label{tab:immerse_codec}
    \begin{tabular}{llccc}
    \toprule
    Eval set & Model & $L1_\theta$ & $L1_\varphi$& $\Delta_{\text{SA}}$ \\
    \midrule
    \multirow{5}{*}{\shortstack[l]{Spatial\\AudioCaps\\(quoted)}}
     & 1D-Conv U-Net (32$\times$)        & 1.55 & 0.29 & 1.56 \\
     & 1D-Conv U-Net (64$\times$)        & 1.57 & 0.24 & 1.57 \\
     & 1D-Conv U-Net (128$\times$)       & 1.57 & 0.16 & 1.57 \\
     & 1D-Conv U-Net-$\beta$ (128$\times$) & 1.43 & 0.15 & 1.51 \\
     & Measurement error (floor)         & 0.17 & 0.14 & 1.03 \\
    \midrule
    \multirow{2}{*}{\shortstack[l]{Spatial\\LibriSpeech(ours)}}
     & Reproduced codec (128$\times$)    & 1.526 & 0.590 & 1.512 \\
     & GT estimator floor                & 0.139 & 0.127 & 0.208 \\
    \midrule
    --- & Chance (target-independent azimuth) & $\pi/2 \approx 1.571$ & -- & -- \\
    \bottomrule
    \end{tabular}
\end{table}

\begin{table}[H]
    \centering
    \caption{
    Oracle reconstruction of ground-truth FOA waveforms (no generation): per-channel Griffin–Lim on ground-truth magnitudes, and BigVGAN per-channel mel round trip. Spatial LibriSpeech test targets; lower is better.}
    \label{tab:bigvgan_recon}
    \setlength{\tabcolsep}{2.5pt}
    \small
    \begin{tabular}{lccccccc}
    \toprule
     & Wave L2 $\downarrow$ & Amp. L2 $\downarrow$ & Phase L2 $\downarrow$ & MRSTFT $\downarrow$ & $L1_\theta$ & $L1_\varphi$& $\Delta_{\text{SA}}$ \\
    \midrule
    Griffin–Lim (GT magnitudes) & 0.013 & 0.007 & 1.553 & 0.552 & 1.567 & 0.402 & 1.560 \\
    BigVGAN reconstruction & 0.035 & 0.544 & 1.552 & 1.575 & 1.607 & 0.384 & 1.606 \\
    GT estimator floor & -- & -- & -- & -- & 0.139 & 0.127 & 0.208 \\
    \midrule
    Chance (target-indep.\ azimuth) & -- & -- & -- & -- & $\pi/2 \approx 1.571$ & -- & -- \\
    \bottomrule
    \end{tabular}
\end{table}

\paragraph{Griffin–Lim on ground-truth magnitudes.}
 The simplest recovery route sets the ceiling most cleanly: we run per-channel Griffin–Lim on the ground-truth target magnitudes (no generation, no magnitude error) and evaluate the reconstruction (Table~\ref{tab:bigvgan_recon}). Magnitude-side fidelity is near-perfect by construction (Wave L2 0.013, Amp. L2 0.007)—matching the best waveform fidelity any model attains in Table~\ref{tab:spatial_main}—yet azimuth sits at the chance level ($L1_\theta = 1.567$ vs. $\pi/2 \approx 1.571$; floor 0.139), while elevation partially survives (0.402 vs. floor 0.127): the same magnitude-recoverable/phase-dependent split as the two routes below, now with all generation and magnitude error removed.

\paragraph{Vocoder reconstruction.}
To isolate the vocoder from generation error, we pass ground-truth FOA waveforms through BigVGAN per channel (mel analysis and re-synthesis, no generation) and evaluate the reconstruction (Table~\ref{tab:bigvgan_recon}). Per-channel fidelity survives the round trip (Wave L2 0.035), but azimuth lands at the chance level ($L1_\theta = 1.607$ vs.\ $\pi/2 \approx 1.571$; floor 0.139), while elevation partially survives (0.384 vs.\ floor 0.127)---the same magnitude-recoverable/phase-dependent split as the codec above. The vocoder route is thus capped before any generative model enters: Tango's adapted variant (Table~\ref{tab:spatial_main}, $L1_\theta = 1.542$) inherits this ceiling rather than failing to learn.

\subsubsection{Seismic Baselines.}
All seismic methods follow the conditioning protocol of HEGGS \citep{jung2025heggs}, whose generator is natively conditioned on the source-station waveform; the GAN and Latent Diffusion Model (LDM) \cite{rombach2022ldm} baselines are adapted to receive the same source conditioning, as done in the HEGGS paper's own baseline comparisons. All methods additionally share the HEGGS conditioning set — source longitude/latitude, station longitude/latitude, back azimuth, source–station distance, source depth, and local magnitude — and the HEGGS conditioning protocol, with preprocessing modified to a wider 0.1–45 Hz bandpass (vs. the 1–45 Hz of \cite{jung2025heggs}) to retain low-frequency content; all else follows their pipeline (60 s three-component windows at 100 Hz).

\paragraph{Latent-diffusion VAE reconstruction (seismic).}

\begin{table}[H]
    \centering
    \caption{
    LDM VAE reconstruction of ground-truth SCEDC waveforms (autoencoding round trip, no generation), reassembled with ground-truth phase vs. per-channel Griffin–Lim recovery.
    }
    \label{tab:vae_recon}
    \setlength{\tabcolsep}{2.5pt}
    \small
    \begin{tabular}{lccccccc|cccccc}
    \toprule
    \multirow{2}{*}{\textbf{Method}}
    & \multicolumn{7}{c|}{\textbf{Goodness-of-fit (GOF)} $\uparrow$}
    & \multicolumn{6}{c}{\textbf{Inter-component particle motion} $\downarrow$} \\
    \cmidrule(lr){2-8} \cmidrule(lr){9-14}
    & PGA & PSA & FAS & Env. & $I_A$ & $D_S$ & Mean
    & PAE & RLE & PAE$_{\mathrm{P}}$ & RLE$_{\mathrm{P}}$ & PAE$_{\mathrm{S}}$ & RLE$_{\mathrm{S}}$ \\
    \midrule
    VAE recon + GT Phase
        & 95.58 & 98.57 & 98.35 & 99.92 & 98.62 & 99.11 & 98.36
        & 4.10 & 0.022 & 2.22 & 0.011 & 3.68 & 0.016 \\
    VAE recon + GL Phase
        & 88.30 & 92.62 & 94.34 & 98.50 & 99.31 & 98.93 & 95.33
        & 41.34 & 0.083 & 28.05 & 0.068 & 42.93 & 0.075 \\
    \bottomrule
    \end{tabular}
\end{table}

The latent route on seismic shows the same ceiling as its spatial counterparts. We pass ground-truth target waveforms through the LDM \cite{rombach2022ldm} baseline's VAE (autoencoding round trip, no generation) and reassemble with either the ground-truth phase or per-channel Griffin–Lim recovery (Table~\ref{tab:vae_recon}). With ground-truth phase, reconstruction is near-perfect on both axes (mean GOF 98.36; PAE$_{\mathrm{S}}$ = $3.7^\circ$): the VAE itself is not the bottleneck. Substituting Griffin–Lim recovery barely moves the amplitude-based scores (mean GOF 95.33, still "excellent" on the Olsen–Mayhew \cite{olsen2010goodness} scale) while degrading S-wave polarization by an order of magnitude ($3.7^\circ$ $\to$ $42.9^\circ$)—the seismic counterpart of the metric insensitivity in Tables~\ref{tab:immerse_codec} and ~\ref{tab:bigvgan_recon}, now with all generation error removed. The split between arrivals repeats the pattern of the spatial routes: P-wave polarization, partially constrained by magnitudes, survives recovery far better ($28.1^\circ$) than S-wave polarization does. Notably, this oracle-magnitude recovery ceiling (PAE$_{\mathrm{S}}$ $42.9^\circ$) is still $9.1^\circ$ worse than RIPPLE operating on its own predicted magnitudes ($33.8^\circ$, Table~\ref{tab:seismic_main}): even perfect magnitudes do not rescue per-channel phase recovery.

\paragraph{Latent-diffusion generation (seismic).}

\begin{table}[H]
    \centering
    \caption{Latent-diffusion (LDM) generation on SCEDC cross-station
    translation, evaluated on the same test targets as Table~\ref{tab:seismic_main}.
    Random-direction PAE: 57.3$^\circ$. GOF higher is better;
    particle-motion errors lower is better.}
    \label{tab:ldm-gen}
    \setlength{\tabcolsep}{2.5pt}
    \small
    \begin{tabular}{lccccccc|cccccc}
    \toprule
    & \multicolumn{7}{c|}{Goodness-of-fit (GOF) $\uparrow$}
    & \multicolumn{6}{c}{Inter-component particle motion $\downarrow$} \\
    Method & PGA & PSA & FAS & Env. & $I_A$ & $D_S$ & Mean
    & PAE & RLE & PAE$_P$ & RLE$_P$ & PAE$_S$ & RLE$_S$ \\
    \midrule
    LDM \cite{rombach2022ldm}
    & 39.46 & 42.70 & 38.47 & 84.68 & 37.07 & 71.05 & 52.24
    & 50.02 & 0.200 & 35.03 & 0.169 & 56.84 & 0.229 \\
    \bottomrule
    \end{tabular}
\end{table}

Table~\ref{tab:vae_recon} isolates the LDM VAE from generation error by round-tripping ground-truth waveforms; here we evaluate the full LDM generator \cite{rombach2022ldm} trained on our cross-station protocol, on the same test targets as Table~\ref{tab:seismic_main} (Table~\ref{tab:ldm-gen}). Its amplitude-based GOF is competitive (mean 52.2, comparable to HEGGS at 49.9), yet its S-wave polarization sits at the random-direction level (PAE$_{\mathrm{S}}$ $56.8^\circ$ vs.\ $57.3^\circ$)—indistinguishable from chance, and no better than HEGGS ($55.0^\circ$) or RIPPLE (GL) ($51.9^\circ$). A third architecture thus reaches the same ceiling: the bottleneck is per-channel phase recovery, not the generative backbone.

\subsection{Implementation details.}\label{app:implementaion_details}
\paragraph{Loss schedule.}
In Eq.~\ref{eq:total-loss} we fix $\lambda_1=\lambda_2=\lambda_5=1$ throughout training, while the phase quality terms are warmed up: $\lambda_3 = \lambda_4 = \min\!\bigl(\max\bigl((e - e_0)/\Delta e,\, 0\bigr),\, 1\bigr)$ at epoch $e$, ramping linearly from 0 to 1 over epochs 5--15 on Spatial LibriSpeech and epochs 150--200 on SCEDC.
Early training thus prioritizes velocity-learning stability; the phase quality and IPD terms take full effect once the velocity fields have stabilized.

\paragraph{Phase quality loss.}
$\Lphaseq = 0.1\,L_{\mathrm{dir}} + 0.5\,L_{\mathrm{IF}} + 0.1\,L_{\mathrm{unit}}$, where (a) $L_{\mathrm{dir}}$ is the magnitude-weighted cosine-similarity loss between $(\hat c,\hat s)$ and $(\cos\angle\Xt,\sin\angle\Xt)$; (b) $L_{\mathrm{IF}}$ is an instantaneous-frequency loss comparing first-order temporal differences on $S^1$ via the cross product $\hat c_t\hat s_{t+1}-\hat s_t\hat c_{t+1}\approx\sin(\phi_{t+1}-\phi_t)$, avoiding explicit phase unwrapping, with an L1 penalty against the same quantity computed from the target; and (c) $L_{\mathrm{unit}}$ is the unit-norm penalty $(\hat c^2+\hat s^2-1)^2$. All three terms are gated by Eq.~\ref{eq:mag_weight}.

\paragraph{Training configuration.}
Both velocity networks are UNet-based; the prior uses $K{=}8$ Griffin--Lim iterations and $\sigma_{\text{prior}}{=}0.1$.
All models are trained with AdamW at learning rate $10^{-5}$, batch size 4, and gradient accumulation over 8 steps (effective batch size 32).
Our model trains for 50 epochs on Spatial LibriSpeech and 300 on SCEDC; all baselines we train or adapt use 300 epochs on their respective domain (Diff-SAGe$^\dagger$ and Tango$^\dagger$ on spatial; HEGGS, GAN, and LDM on seismic).
Spatial experiments run on two NVIDIA A100 80GB GPUs, seismic on two RTX A6000s.

\subsection{Evaluation metrics.}\label{app:metrics}
\paragraph{Goodness-of-fit (GOF).}
Per-component scores are $\text{GOF} = 100 \cdot \mathrm{erfc}\big(2|x-y|/(x+y)\big)$ \citep{olsen2010goodness}, computed on six intensity measures: peak ground acceleration (PGA); pseudo-spectral acceleration (PSA, 0.1--10\,s, 5\% damped); Konno--Ohmachi-smoothed Fourier amplitude spectrum (FAS); Arias intensity ($I_A$); significant duration ($D_{S5\text{--}95}$); and envelope correlation (Env.).

\paragraph{Polarization metrics.}
PAE/RLE are computed in sliding windows over the same evaluation interval as the GOF measures: in each window we eigendecompose the $3{\times}3$ covariance of the three components; the principal eigenvector gives the particle-motion direction (compared via the angle between target and prediction, with sign ambiguity resolved by $|\cos|$) and the eigenvalue ratios give rectilinearity.
This is the same analysis practitioners run on real three-component records, so PAE measures usability for the workflows that cross-station translation is meant to serve.
Window contributions are weighted by target energy and target rectilinearity, since polarization direction is only well-defined for energetic, rectilinear motion---the same rationale as our magnitude-weighted phase loss.
PAE$_{\mathrm{P}}$/PAE$_{\mathrm{S}}$ use fixed 2\,s windows anchored at the TauP-predicted arrivals.
Two random directions yield an expected PAE of $57.3^\circ$.

To support the ``near chance'' characterization, we compute 95\% bootstrap confidence intervals for PAE$_\mathrm{S}$ (10{,}000 resamples over the 861 test events): GAN $[56.7, 61.4]$ and HEGGS $[52.7, 57.3]$ both include the $57.3^\circ$ random-direction expectation, i.e., their S-wave polarization is statistically indistinguishable from chance; RIPPLE (GL) $[49.5, 54.2]$ sits slightly but significantly below it; RIPPLE $[30.8, 36.9]$ is separated from chance by over $20^\circ$.

\section{Ablation Results}\label{app:ablations}

Tables~\ref{tab:ablation_spatial_full} and~\ref{tab:ablation_seismic_full} extend the two domains of Table~\ref{tab:ablation_merged} to all metrics. The behavior of the omitted metrics supports the column selection in the main text.

\begin{table*}[h]
    \centering
    \caption{Full ablation on Spatial LibriSpeech (cross-room), extending the spatial columns of Table~\ref{tab:ablation_merged}. Each row removes one component from \emph{RIPPLE} while keeping the magnitude path and all other settings fixed. \emph{w/o source-coherent init} replaces $\phi_{\text{init}} = \angle X^{(\mathrm{s})}$ with random phase initialization while keeping the Griffin--Lim prior; \emph{w/o GL prior} replaces $p_0$ with a Gaussian prior; \emph{w/o IPD} sets $\lambda_4{=}0$; \emph{coupled timesteps} ties $t_m{=}t_p$. Lower is better.}
    \label{tab:ablation_spatial_full}
    \setlength{\tabcolsep}{4pt}
    \renewcommand{\arraystretch}{1.15}
    \begin{tabular}{lccccccc}
    \toprule
    & \multicolumn{4}{c}{Fidelity} & \multicolumn{3}{c}{Spatial accuracy} \\
    \cmidrule(lr){2-5}\cmidrule(lr){6-8}
    Configuration & Wave L2 $\downarrow$ & Amp.\ L2 $\downarrow$ & Phase L2 $\downarrow$ & MRSTFT $\downarrow$ & $L1_\theta$ $\downarrow$ & $L1_\varphi$ $\downarrow$ & $\Delta_{\text{Spatial-Angle}}$ $\downarrow$ \\
    \midrule
    \textbf{RIPPLE} & 0.019 & \textbf{0.432} & \textbf{0.619} & \textbf{1.249} & \textbf{0.243} & \textbf{0.152} & \textbf{0.319} \\
    w/o source-coherent init & \textbf{0.018} & 0.448 & 0.683 & 1.234 & 0.253 & 0.161 & 0.332 \\
    w/o Griffin--Lim prior & 0.019 & 0.581 & 1.404 & 1.413 & 1.616 & 0.615 & 1.604 \\
    w/o IPD loss ($\lambda_4{=}0$) & 0.022 & 0.502 & 1.451 & 1.283 & 1.627 & 0.188 & 1.625 \\
    coupled timesteps ($t_m{=}t_p$) & 0.021 & 0.557 & 0.807 & 1.432 & 1.582 & 0.200 & 1.581 \\
    \midrule
    RIPPLE (GL) & 0.022 & 0.460 & 1.072 & 1.318 & 1.591 & 0.395 & 1.586 \\
    \bottomrule
    \end{tabular}
\end{table*}

\paragraph{Spatial (Table~\ref{tab:ablation_spatial_full}).}
Waveform-level fidelity is essentially insensitive to the phase-side configuration: Wave L2 stays within $0.018$--$0.022$ across all rows, showing that with the magnitude path fixed, the time-domain envelope is preserved even under phase collapse. In contrast, Phase L2 and the angular metrics separate by up to a fivefold regression across configurations; this contrast itself reconfirms, at the metric level, the premise that spatial information is carried by phase alignment rather than amplitude. The coupled-timestep row shows the separation most sharply: Phase L2 degrades mildly ($0.807$) while $L1_\theta$ collapses ($1.582$), localizing the damage of timestep coupling to inter-channel alignment rather than per-bin phase accuracy.

\begin{table*}[h]
    \centering
    \caption{Full ablation on SCEDC (cross-station), extending the seismic columns of Table~\ref{tab:ablation_merged}. GOF scores in $[0,100]$, higher is better; particle-motion errors, lower is better (random-direction baseline: PAE $= 57.3^\circ$). Intensity-measure definitions in Appendix~\ref{app:metrics}. All configurations receive identical conditioning, including the source-station waveform.}
    \label{tab:ablation_seismic_full}
    \setlength{\tabcolsep}{3.5pt}
    \renewcommand{\arraystretch}{1.15}
    \resizebox{0.99\textwidth}{!}{
    \begin{tabular}{lccccccc|cccccc}
    \toprule
    & \multicolumn{7}{c|}{Goodness-of-fit (GOF) $\uparrow$} & \multicolumn{6}{c}{Inter-component particle motion $\downarrow$} \\
    Configuration & PGA & PSA & FAS & Env. & $I_A$ & $D_S$ & Mean & PAE & RLE & PAE$_\mathrm{P}$ & RLE$_\mathrm{P}$ & PAE$_\mathrm{S}$ & RLE$_\mathrm{S}$ \\
    \midrule
    \textbf{RIPPLE} & \textbf{90.56} & \textbf{55.86} & 47.01 & \textbf{85.75} & \textbf{67.73} & \textbf{65.85} & \textbf{68.79} & \textbf{32.57} & \textbf{0.111} & \textbf{26.48} & \textbf{0.121} & \textbf{33.81} & \textbf{0.098} \\
    w/o source-coherent init & 87.15 & 52.49 & 47.13 & 85.41 & 55.31 & 63.33 & 65.13 & 43.28 & 0.142 & 33.17 & 0.169 & 48.23 & 0.110 \\
    w/o Griffin--Lim prior & 67.03 & 52.58 & 47.45 & 85.05 & 39.26 & 64.97 & 59.39 & 52.07 & 0.167 & 36.54 & 0.187 & 57.33 & 0.170 \\
    w/o IPD loss ($\lambda_4{=}0$) & 70.38 & 53.24 & 46.92 & 84.49 & 41.10 & 64.11 & 60.04 & 52.38 & 0.171 & 35.42 & 0.183 & 57.31 & 0.170 \\
    coupled timesteps ($t_m{=}t_p$) & 72.98 & 50.99 & 45.31 & 85.12 & 38.44 & 54.2 & 57.84 & 50.98 & 0.135 & 34.33 & 0.170 & 50.28 & 0.115 \\
    \midrule
    RIPPLE (GL) & 81.96 & 53.05 & \textbf{47.61} & 85.61 & 41.20 & 65.64 & 62.51 & 49.79 & 0.169 & 40.42 & 0.181 & 51.87 & 0.153 \\
    \bottomrule
    \end{tabular}
    }
\end{table*}

\paragraph{Seismic (Table~\ref{tab:ablation_seismic_full}).}
The six GOF measures split into two groups by their sensitivity to phase learning. FAS ($46.9$--$47.6$), envelope correlation ($84.5$--$85.8$), and significant duration ($63.3$--$65.9$) are essentially invariant across all configurations---purely spectral or low-frequency envelope-level quantities determined by the magnitude path alone; this is the full evidence behind the main-text observation that purely spectral quantities are unchanged. In contrast, PGA ($67.0$--$90.6$) and Arias intensity ($39.3$--$67.7$), both sensitive to time-domain reassembly, separate widely. The same split appears in the learned-phase gains over the Griffin--Lim ablation: the improvement concentrates on time-domain intensity measures sensitive to waveform reassembly ($I_A$ $41.2 \to 67.7$, PGA $82.0 \to 90.6$), while purely spectral quantities are unchanged (FAS $47.6$ vs.\ $47.0$). Notably, under \emph{w/o source-coherent init}, PGA drops only mildly ($87.2$) while $I_A$ regresses substantially ($67.7 \to 55.3$): accumulating energy over the full waveform ($I_A$) demands sustained phase alignment in a way that an instantaneous peak (PGA) does not. The RLE family moves in the same direction as PAE throughout---RLE$_\mathrm{S}$ converges to an identical $0.170$ under both \emph{w/o GL} and \emph{w/o IPD}---confirming that polarization direction and rectilinearity are two views of the same cross-component structure, which is why the main text reports PAE only.

\subsection{Sensitivity to prior iterations $K$.}\label{app:k_sensitivity}

\begin{figure}[t]
    \centering
    \includegraphics[width=0.5\linewidth]{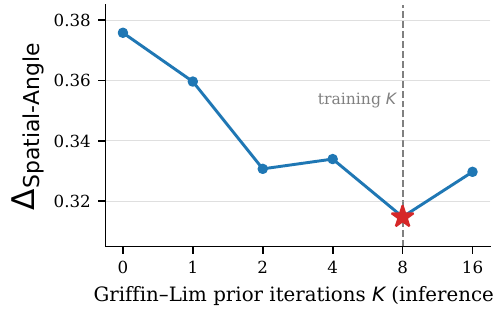}
    \caption{Sensitivity of $\Delta_{\text{Spatial-Angle}}$ to the number of Griffin--Lim prior iterations $K$ at inference (same 100 test pairs as Figure~4, integrated with $N{=}50$ Heun steps). The flow is trained with $K{=}8$ (dashed line). Lower is better.}
    \label{fig:k_sensitivity}
\end{figure}

The number of Griffin--Lim iterations $K$ controls the quality of the prior: larger $K$ improves magnitude consistency but, since the iteration is per-channel, gradually attenuates the source's inter-channel structure. Since retraining per $K$ is beyond our compute budget, we train with a fixed $K{=}8$ and vary $K$ at inference only; the resulting curve (Figure~\ref{fig:k_sensitivity}) therefore measures sensitivity to prior quality under a fixed flow, conflating prior degradation with train--test prior mismatch. 
$\Delta_{\text{Spatial-Angle}}$ stays within a narrow band ($0.315$--$0.376$ for $K \in [0,16]$)---a spread an order of magnitude smaller than the gap to the Gaussian-prior level ($\approx\!1.5$; Figure~\ref{fig:nfe_curve})—and even $K{=}0$, the source phase with no magnitude projection, remains far below that level.
The trained flow is thus insensitive to the exact prior quality it is integrated from: for this model, the choice of prior (Figure~\ref{fig:nfe_curve} and Table~\ref{tab:ablation_merged}) is first-order, while inference-time $K$ is not a sensitive hyperparameter; whether the same holds for flows trained at other $K$ remains untested.

The $K=0$ endpoint deserves separate comment. There the prior is the source phase with no magnitude projection applied: a valid point of $\mathbb{S}^1$ carrying the source's inter-channel structure, but not magnitude-consistent. This is the projection-free prior identified in the Discussion as separating the two properties that the Gaussian ablation removes jointly. Its $\Delta_{\text{Spatial-Angle}}$ of 0.376 sits far closer to the full model (0.319) than to the Gaussian prior (1.604), indicating that valid $\mathbb{S}^1$ structure together with source coherence---rather than magnitude consistency---accounts for most of the prior's contribution. The evidence is partial: $K$ is varied at inference under a flow trained with $K=8$, so prior degradation is confounded with train--test mismatch.

\section{Training and Inference Algorithms}
\label{app:algorithms}
\subsection{Training and Inference Details}
\label{app:algorithms}

Algorithms~\ref{alg:train} and~\ref{alg:inference} give pseudocode for the training step and the two-stage inference procedure. Here we expand the integration scheme summarized in the main text.

\paragraph{Magnitude stage.}
We integrate the magnitude flow from $\log|\Xs|$ to a predicted target magnitude $\log|\hat\Xt|$ with a Heun second-order scheme along the uniform timestep grid $\{t_i\}_{i=0}^{N}$ ($t_i = i/N$), and set $\hat m = \exp(z)$.

\paragraph{Prior construction.}
From the \emph{predicted} target magnitude $\hat m$ and the source phase $\angle\Xs$ we construct the source-informed Griffin--Lim prior $p_0$ (Eq.~\ref{eq:p0}). This mirrors the training-time prior, except that training uses the oracle target magnitude $|\Xt|$ while inference uses the predicted $\hat m$.

\paragraph{Phase stage.}
We integrate the phase ODE from $p_0$ using the same Heun scheme. At every phase step, the predicted log-magnitude $z$ is concatenated along the channel dimension as conditioning, matching the training input format. After integration we project the phase prediction onto $\Sone$ by unit normalization, reassemble the complex STFT $\hat X = \hat m \odot (\hat\cos + i\,\hat\sin)$, and recover the waveform by inverse STFT. We use $N=50$ as a default.

\subsection{Geometry of the phase flow}\label{app:p_0}

The phase flow is defined on $\mathbb{S}^1$ but integrated in $\mathbb{R}^2$. The interpolant of Eq.~\ref{eq:phase_inter} traces the chord between $p_0$ and $(\cos\angle X^{(t)}, \sin\angle X^{(t)})$, so for $t_p \in (0,1)$ the training state lies strictly inside the unit circle: the parameterization is ambient by construction, not only at its endpoints.

This makes the perturbation $\sigma_{\text{prior}}\epsilon$ in Eq.~\ref{eq:p0} structural rather than incidental. With $\sigma_{\text{prior}} = 0$, training states would concentrate on the thin family of chords joining prior to target, leaving the velocity field unconstrained in the radial direction---while integration departs from that family immediately, since the Heun predictor is a straight extrapolation in $\mathbb{R}^2$ and the corrector averages two velocities evaluated at different radii. The network would then be queried away from the states it was trained on. The perturbation thickens the support into a neighborhood of the chords, and the two mechanisms that return predictions to the circle act at separate points: the unit-norm penalty in $\mathcal{L}_{\text{phase}}^q$ pulls the one-step prediction toward $\mathbb{S}^1$ during training, and the final unit normalization (Appendix~\ref{app:algorithms}) enforces it exactly at inference.

The scale is bounded above by what it is meant to leave intact: $\sigma_{\text{prior}}$ must remain small relative to the inter-channel structure carried by $p_0$, since a large perturbation destroys precisely the coherence that source-informed initialization supplies. We fix $\sigma_{\text{prior}} = 0.1$ throughout and do not tune it; sensitivity to this choice is untested, note also that $\sigma_{\text{prior}}$ varies prior quality along a different axis than the $K$ sweep of Appendix~\ref{app:k_sensitivity}.

\paragraph{Why not a manifold-native flow.} Riemannian flow matching defines interpolant and velocity on the manifold and would remove the ambient excursion above. We keep the embedded $\mathbb{R}^2$ parameterization for two reasons. First, it leaves the phase head architecturally identical to the magnitude head, so the two velocity networks differ only in input channels and share the conditioning encoder $E_c$ without map-specific machinery. Second, the predicted log-magnitude is concatenated to the phase head's input at every integration step; in the ambient parameterization this is a plain channel concatenation, whereas a manifold-native flow would place state and conditioning in different spaces. What the ambient view loses is restored where it matters: the IPD loss (Eq~\ref{eq:ipd_loss}) normalizes predictions onto $\mathbb{S}^1$ before forming relative phase vectors, so the coherence objective is evaluated on the manifold regardless of where the flow is integrated.

\begin{minipage}[t]{0.45\textwidth}
\vspace{0pt}
    \begin{algorithm}[H]
        \caption{Training step}
        \label{alg:train}
        \begin{algorithmic}[1]
            \REQUIRE paired sample $(\ws, \wt)$, conditioning $c$,
                     loss weights $\lambda_1,\dots,\lambda_5$
            \STATE $X^{(s)}\gets\Fop\{\ws\}$, $X^{(t)}\gets\Fop\{\wt\}$
            \STATE \textit{// Source-informed phase prior}
            \STATE $\phi_{\GL}\gets\GL_K\!\big(|\Xt|;\,\angle\Xs\big)$
            \STATE $p_0\gets(\cos\phi_{\GL},\sin\phi_{\GL})+\sigma_{\text{prior}}\epsilon$,
                   \ $\epsilon\sim\mathcal{N}(0,I)$
            \STATE \textit{// Independent timesteps}
            \STATE Sample $t_m, t_p \sim \mathcal{U}(0,1)$ independently
            \STATE \textit{// Interpolation and target velocities}
            \STATE $z^{\text{mag}}_{t_m} \gets (1{-}t_m)\log|\Xs| + t_m\log|\Xt|$ 
            \STATE $z^{\text{phase}}_{t_p} \gets (1{-}t_p)\,p_0
                   + t_p\,(\cos\angle\Xt, \sin\angle\Xt)$
            \STATE $v^{*}_{\text{mag}}\gets\log|\Xt|-\log|\Xs|$
            \STATE $v^{*}_{\text{phase}}\gets(\cos\angle\Xt,\sin\angle\Xt)-p_0$
            \STATE \textit{// Velocity predictions}
            \STATE $h\gets E_c(c)$
            \STATE $v^{\text{mag}}_\theta\gets v^{\text{mag}}_\theta\!\big(z^{\text{mag}}_{t_m},\,t_m,\,h\big)$
            \STATE $v^{\text{phase}}_\theta\gets v^{\text{phase}}_\theta\!\big([z^{\text{phase}}_{t_p},\,\log|\Xt|],\,t_p,\,h\big)$
                   \COMMENT{phase head concats target magnitude}
            \STATE \textit{// One-step predictions for auxiliary losses}
            \STATE $\hat z^{\text{mag}}_{1|t_m}\gets z^{\text{mag}}_{t_m}+(1{-}t_m)\,v^{\text{mag}}_\theta$
            \STATE $\hat z^{\text{phase}}_{1|t_p}\gets z^{\text{phase}}_{t_p}+(1{-}t_p)\,v^{\text{phase}}_\theta$
            \STATE \textit{// Compose loss (Eq.\ \ref{eq:total-loss})}
            \STATE $\mathcal{L}\gets\lambda_1\Lmag+\lambda_2\Lphasevel
                   +\lambda_3\Lphaseq+\lambda_4\Lipd+\lambda_5\Lrec$
            \STATE Update $\theta$ by a gradient step on $\mathcal{L}$
        \end{algorithmic}
    \end{algorithm}
\end{minipage}
\hfill
\begin{minipage}[t]{0.49\textwidth}
\vspace{0pt}
    \begin{algorithm}[H]
        \caption{Two-stage inference}
        \label{alg:inference}
        \begin{algorithmic}[1]
        \REQUIRE source magnitude $\log|\Xs|$, source phase $\angle\Xs$,
                 conditioning $c$, number of function evaluations $N$
        \ENSURE  target waveform $\hat w^{(t)}$
        \STATE $h \gets E_c(c)$
        \STATE $t_i \gets i/N$ for $i=0,\dots,N$
        \STATE \textit{// Stage 1: magnitude integration ($\log|\Xs|\!\to\!\log|\hat\Xt|$)}
        \STATE $z \gets \log|\Xs|$
        \FOR{$i = 0,\dots,N-1$}
            \STATE $dt \gets t_{i+1}-t_i$
            \STATE $v \gets v^{\text{mag}}_\theta(z, t_i, h)$
            \STATE $z' \gets z + dt\cdot v$ \hfill \COMMENT{predictor}
            \STATE $v' \gets v^{\text{mag}}_\theta(z', t_{i+1}, h)$
            \STATE $z \gets z + \tfrac{1}{2}\,dt\,(v + v')$ \hfill \COMMENT{Heun corrector}
        \ENDFOR
        \STATE $\hat m \gets \exp(z)$
        \STATE \textit{// Source-informed phase prior (input to Stage 2)}
        \STATE $\phi_{\GL} \gets \GL_K\!\big(\hat m;\, \angle\Xs\big)$
        \STATE $p \gets (\cos\phi_{\GL},\sin\phi_{\GL}) + \sigma_{\text{prior}}\,\epsilon$,
               \ $\epsilon\sim\mathcal{N}(0,I)$
        \STATE \textit{// Stage 2: phase integration ($p_0\!\to\!\hat\phi$)}
        \FOR{$i = 0,\dots,N-1$}
            \STATE $dt \gets t_{i+1}-t_i$
            \STATE $v \gets v^{\text{phase}}_\theta\big([p, z],\, t_i,\, h\big)$
            \STATE $p' \gets p + dt\cdot v$
            \STATE $v' \gets v^{\text{phase}}_\theta\big([p', z],\, t_{i+1},\, h\big)$
            \STATE $p \gets p + \tfrac{1}{2}\,dt\,(v + v')$
        \ENDFOR
        \STATE \textit{// Final assembly}
        \STATE $p \gets p / \|p\|_2$ \hfill \COMMENT{project onto $\Sone$}
        \STATE $\hat X \gets \hat m\odot (p_{\cos} + i\, p_{\sin})$
        \STATE $\hat w^{(t)} \gets \Fop^{-1}\!\{\hat X\}$
        \RETURN $\hat w^{(t)}$
        \end{algorithmic}
    \end{algorithm}
\end{minipage}

\section{Additional Qualitative Results}\label{app:add_visual}

\subsection{Spatial Audio}\label{app:spatial_results}

\begin{figure}[H]
    \centering
    \includegraphics[width=0.7\textwidth]{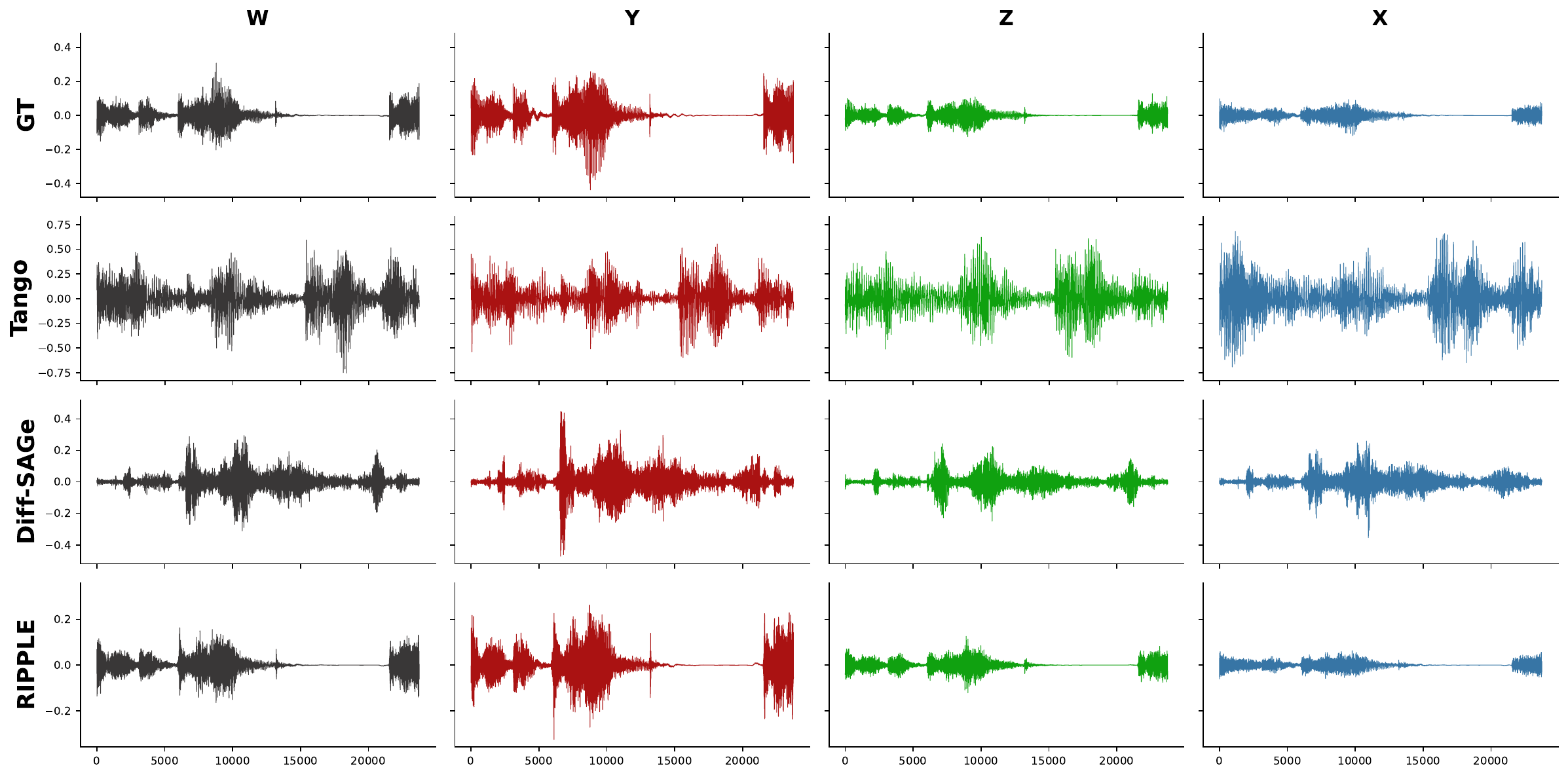}
    \includegraphics[width=0.7\textwidth]{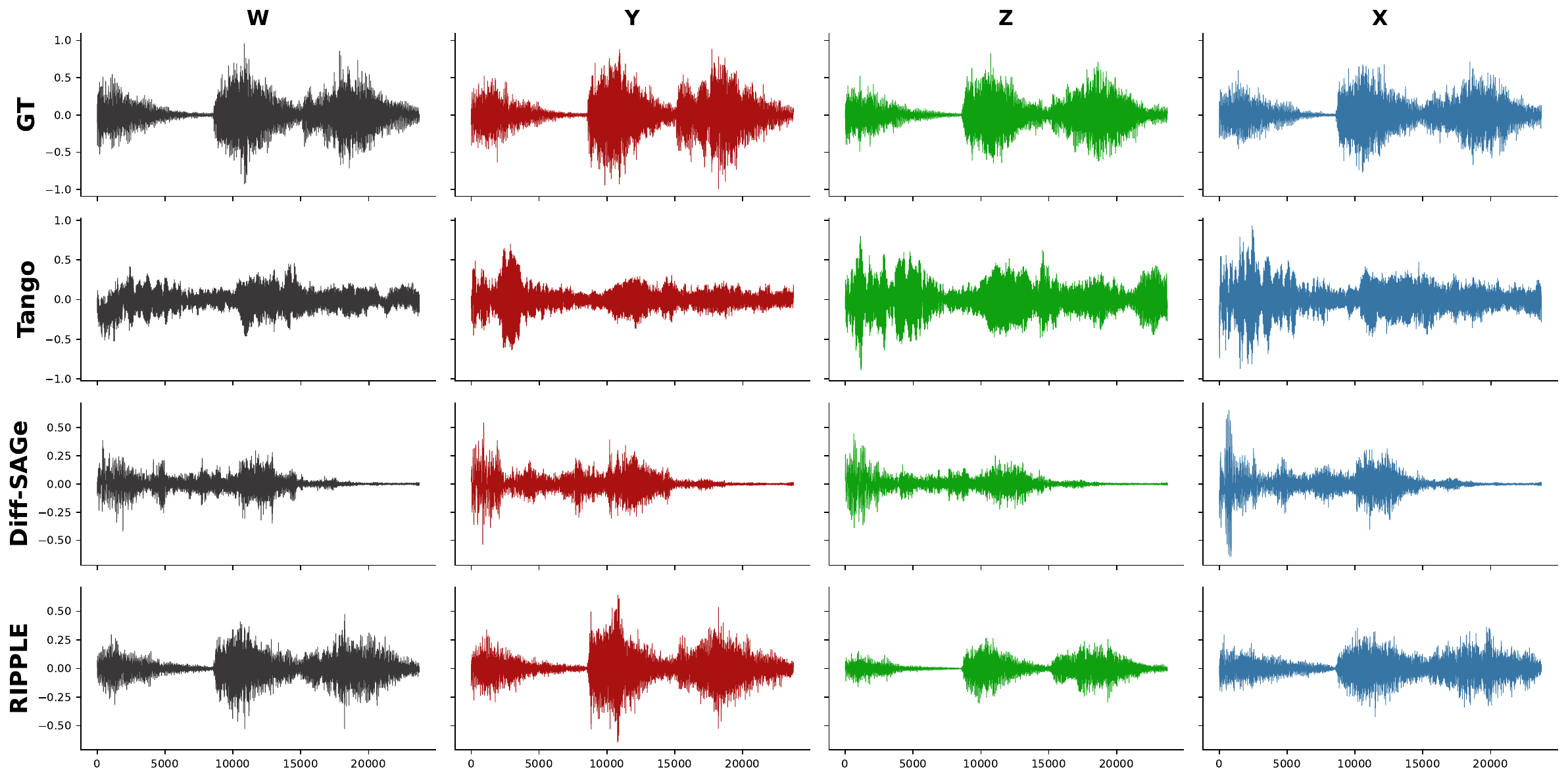}
    \includegraphics[width=0.7\textwidth]{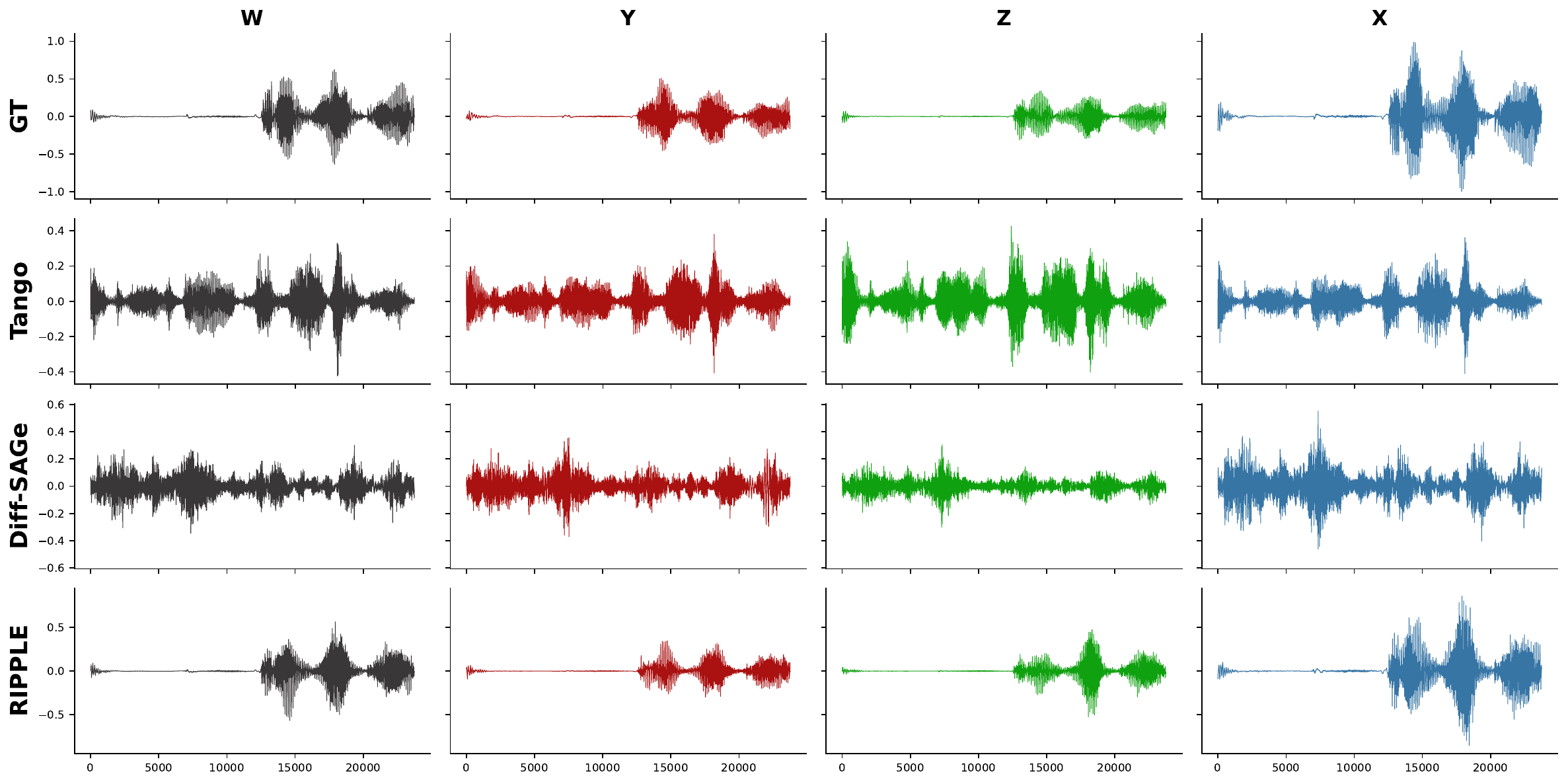}
    \caption{No Source-Target Adaptation}
    \label{fig:app_spatial_direct}
\end{figure}

\begin{figure}[H]
    \centering
    \includegraphics[width=0.7\textwidth]{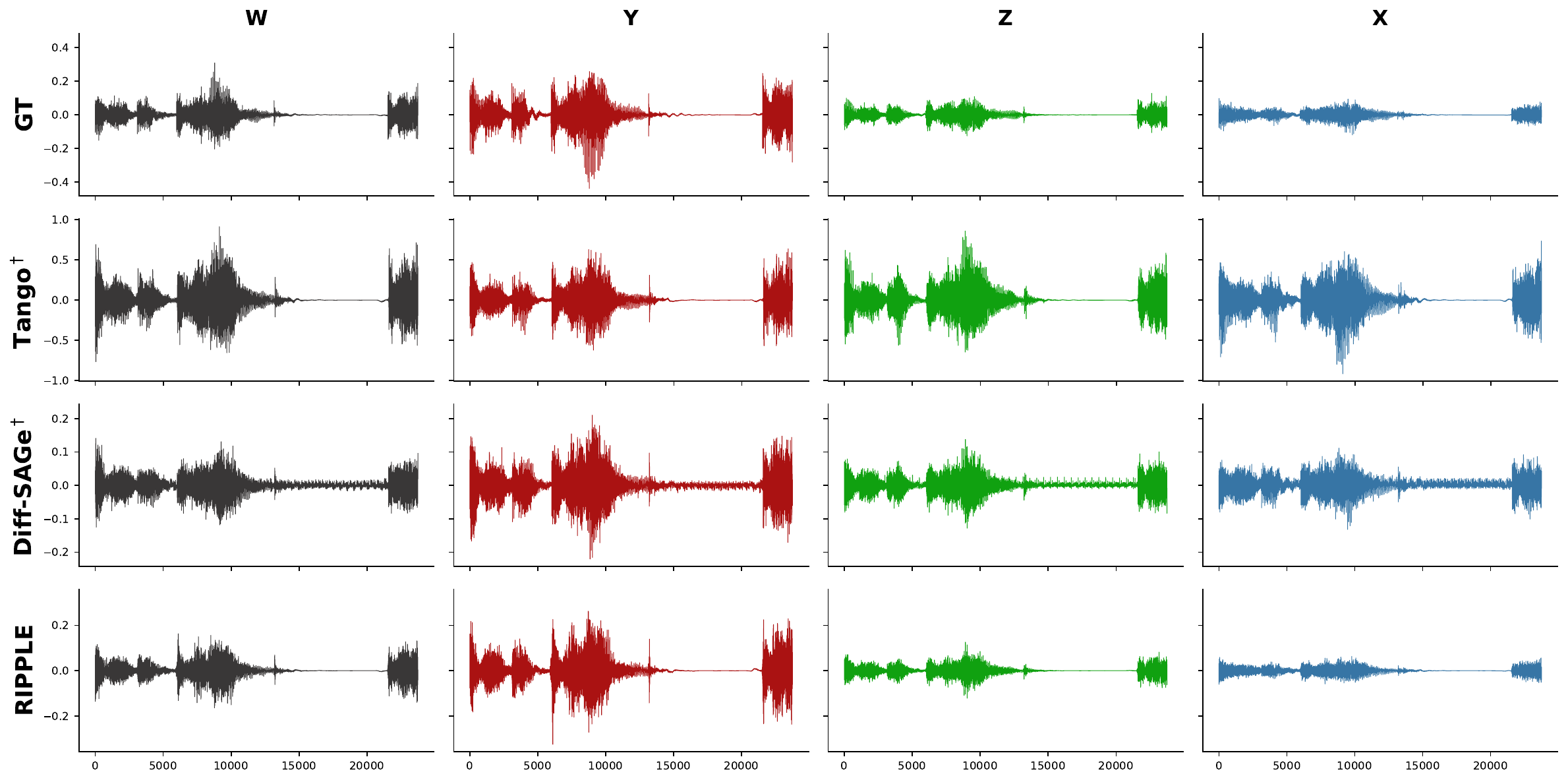}
    \includegraphics[width=0.7\textwidth]{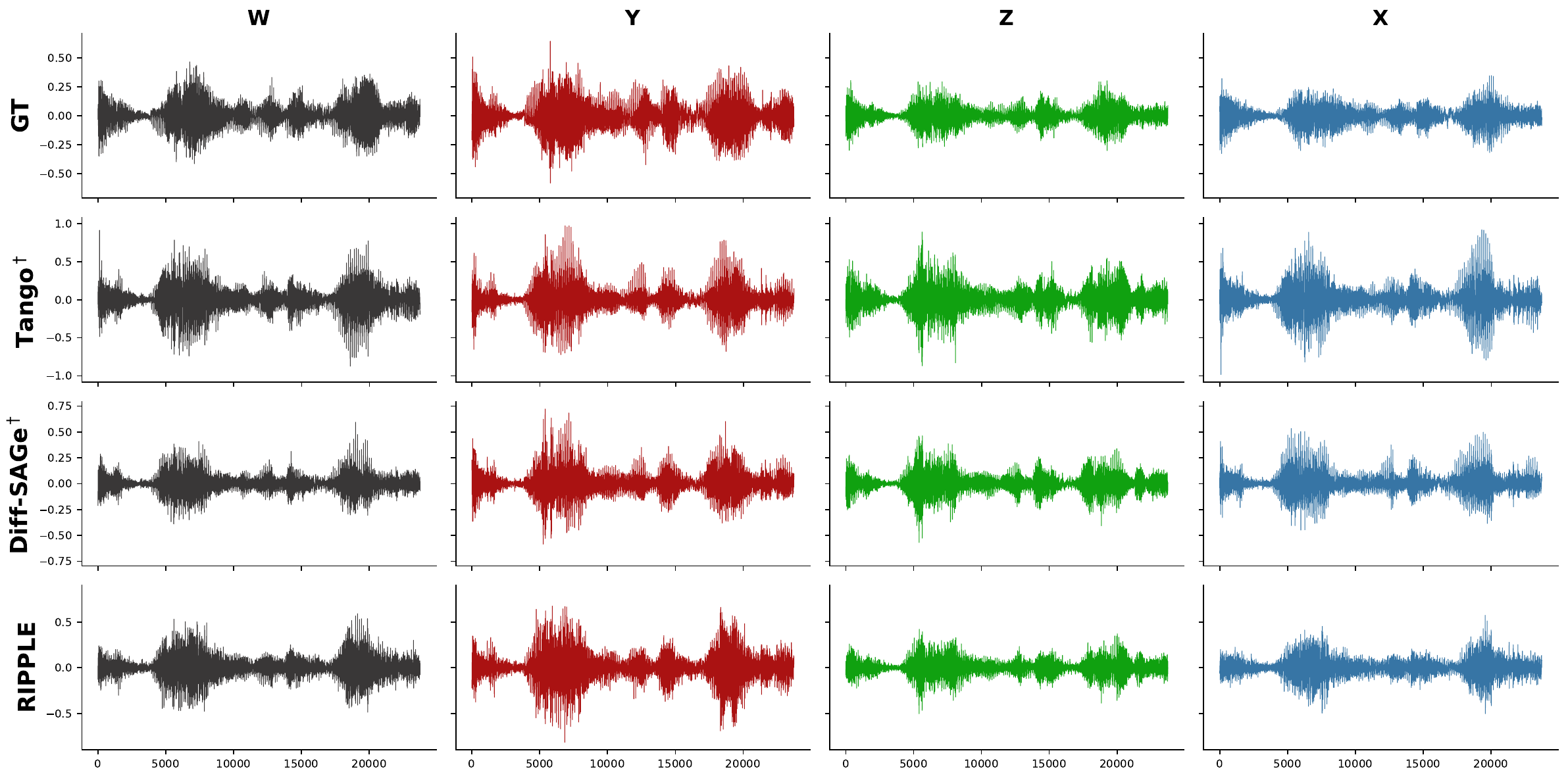}
    \includegraphics[width=0.7\textwidth]{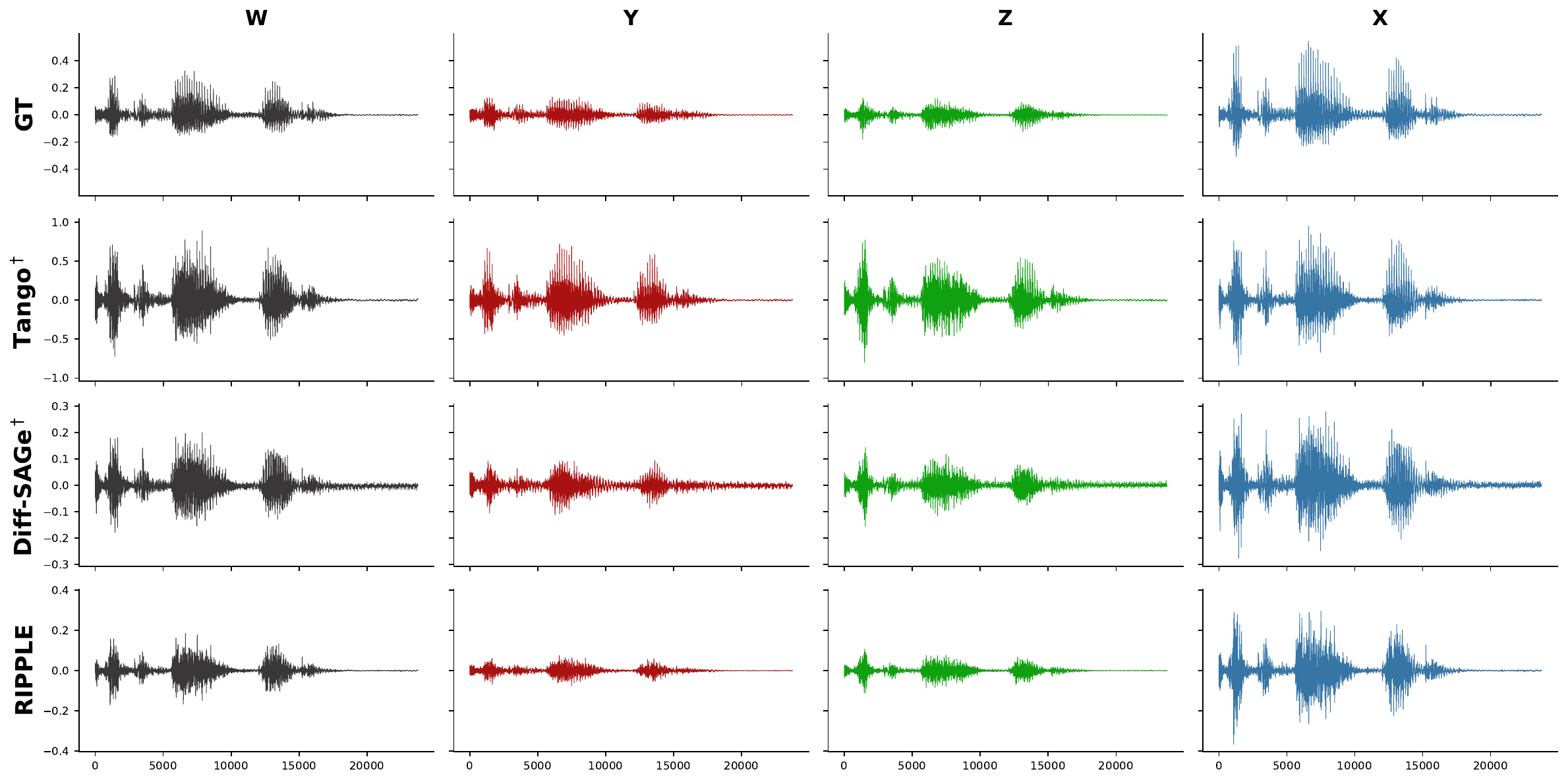}
    \caption{Source-Target Adaptation}
    \label{fig:app_spatial_adapted}
\end{figure}

Figures~\ref{fig:app_spatial_direct} and~\ref{fig:app_spatial_adapted} extend Figure~\ref{fig:waveform_comparison} of the main text to three additional Spatial LibriSpeech test samples each, under the same visualization protocol: four-channel FOA waveforms (W/Y/Z/X) with a shared amplitude scale within each row.

\paragraph{Direct baselines (Figure~\ref{fig:app_spatial_direct}).}
The failure mode of the direct baselines is consistent across samples. Without source conditioning, Tango and Diff-SAGe generate plausible speech-like waveforms, but neither the temporal envelope nor the energy balance between the omnidirectional (W) and directional (Y/Z/X) channels corresponds to the target: the outputs are draws from the marginal distribution rather than translations of this particular scene. This per-sample mismatch is the qualitative counterpart of their fidelity gap in Table~\ref{tab:spatial_main} and motivates the source-adapted comparison below. RIPPLE reproduces both the envelope and the inter-channel amplitude structure in all three samples.

\paragraph{Source-adapted baselines (Figure~\ref{fig:app_spatial_adapted}).}
With source conditioning, both adapted baselines recover the coarse envelope, and each then fails in the manner quantified in Table~\ref{tab:spatial_main}. Tango$^\dagger$ overshoots the overall amplitude scale and concentrates energy on W rather than on the directional channels that dominate the targets; Diff-SAGe$^\dagger$ preserves the scale but attenuates the directional channels, flattening the inter-channel balance that encodes source direction. RIPPLE matches both the scale and the balance across all samples, consistent with its Phase L2 and $L1_\theta$ margins in Table~\ref{tab:spatial_main}.

\subsection{Seismic}\label{app:seismic_results}

\begin{figure}[H]
    \centering
    \includegraphics[width=0.6\textwidth]{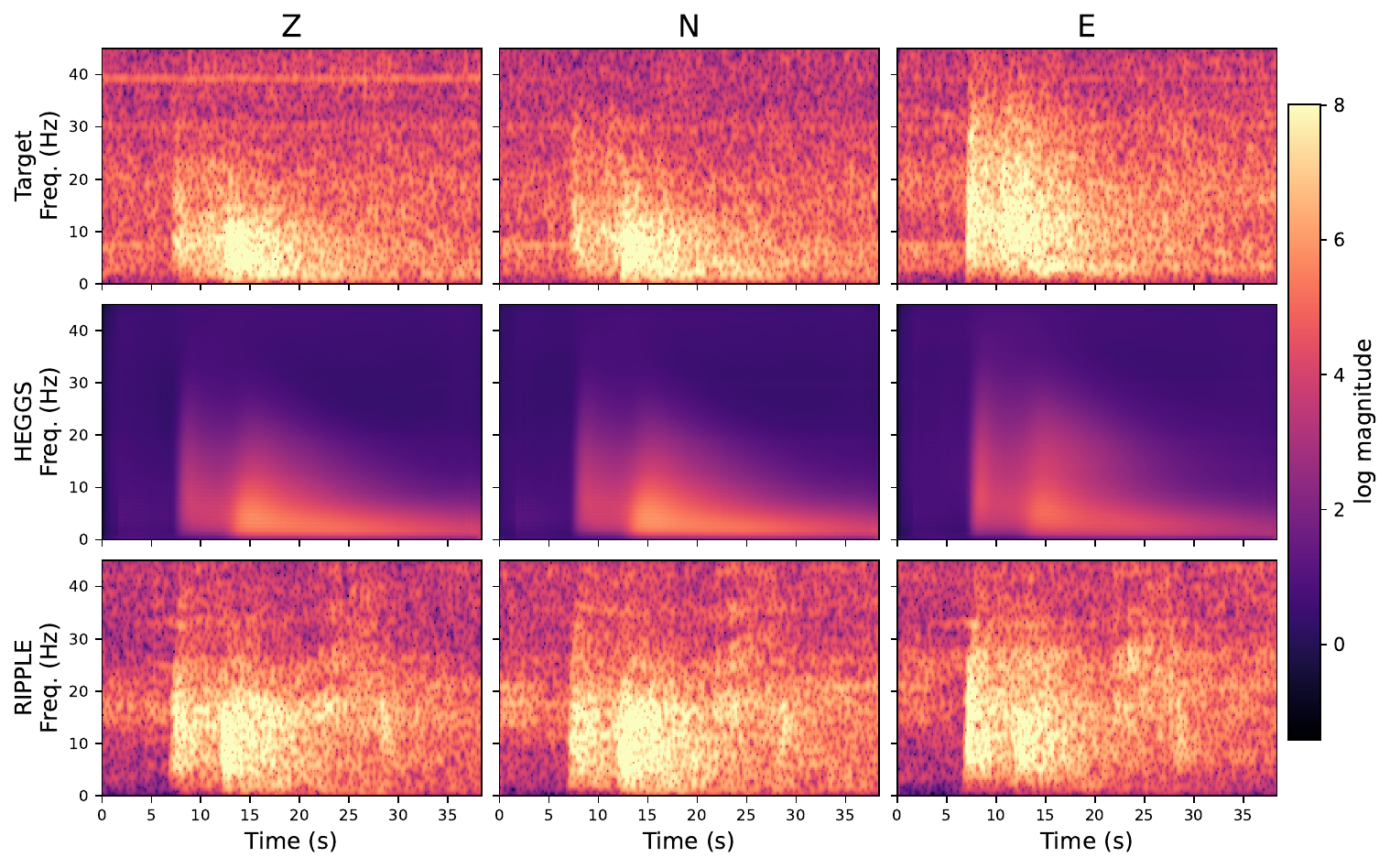}
    \includegraphics[width=0.6\textwidth]{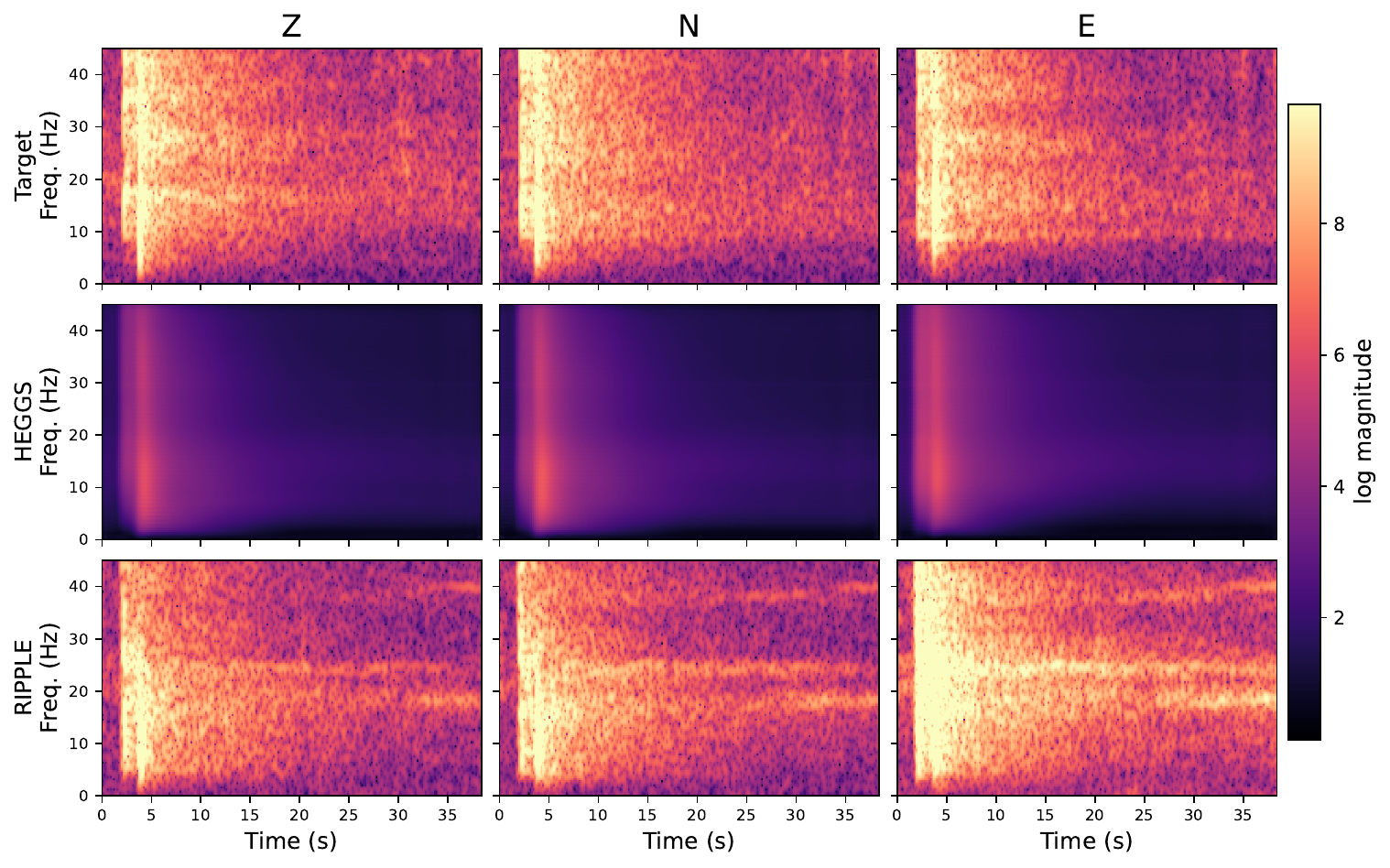}
    \includegraphics[width=0.6\textwidth]{main_seismic/seismic_spec_comparison_182.pdf}
    \caption{Three-component (Z/N/E) log-magnitude spectrograms for three SCEDC cross-station test samples.}
    \label{fig:app_seismic_spec}
\end{figure}

\begin{figure}[H]
    \centering
    \includegraphics[width=0.6\textwidth]{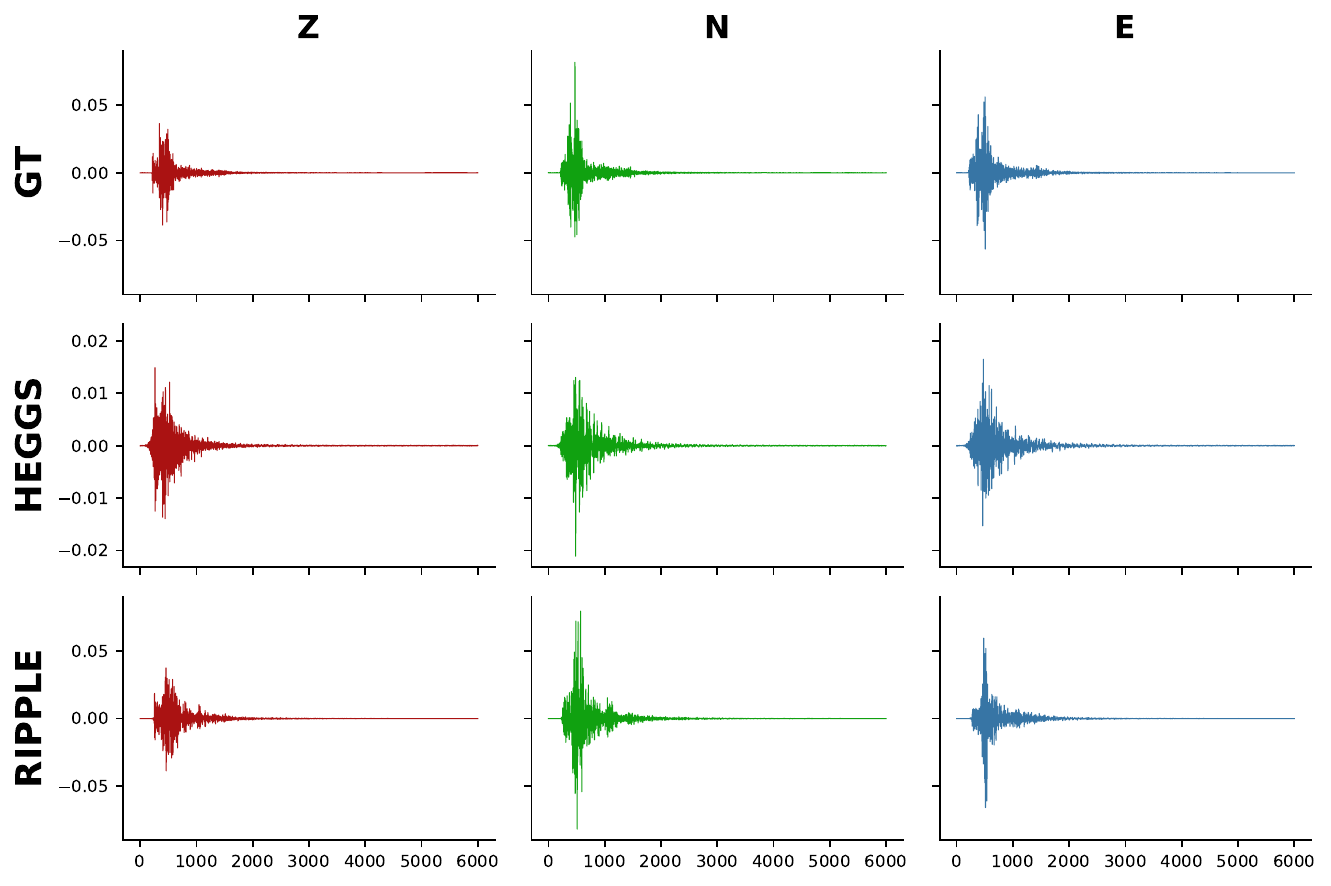}
    \includegraphics[width=0.6\textwidth]{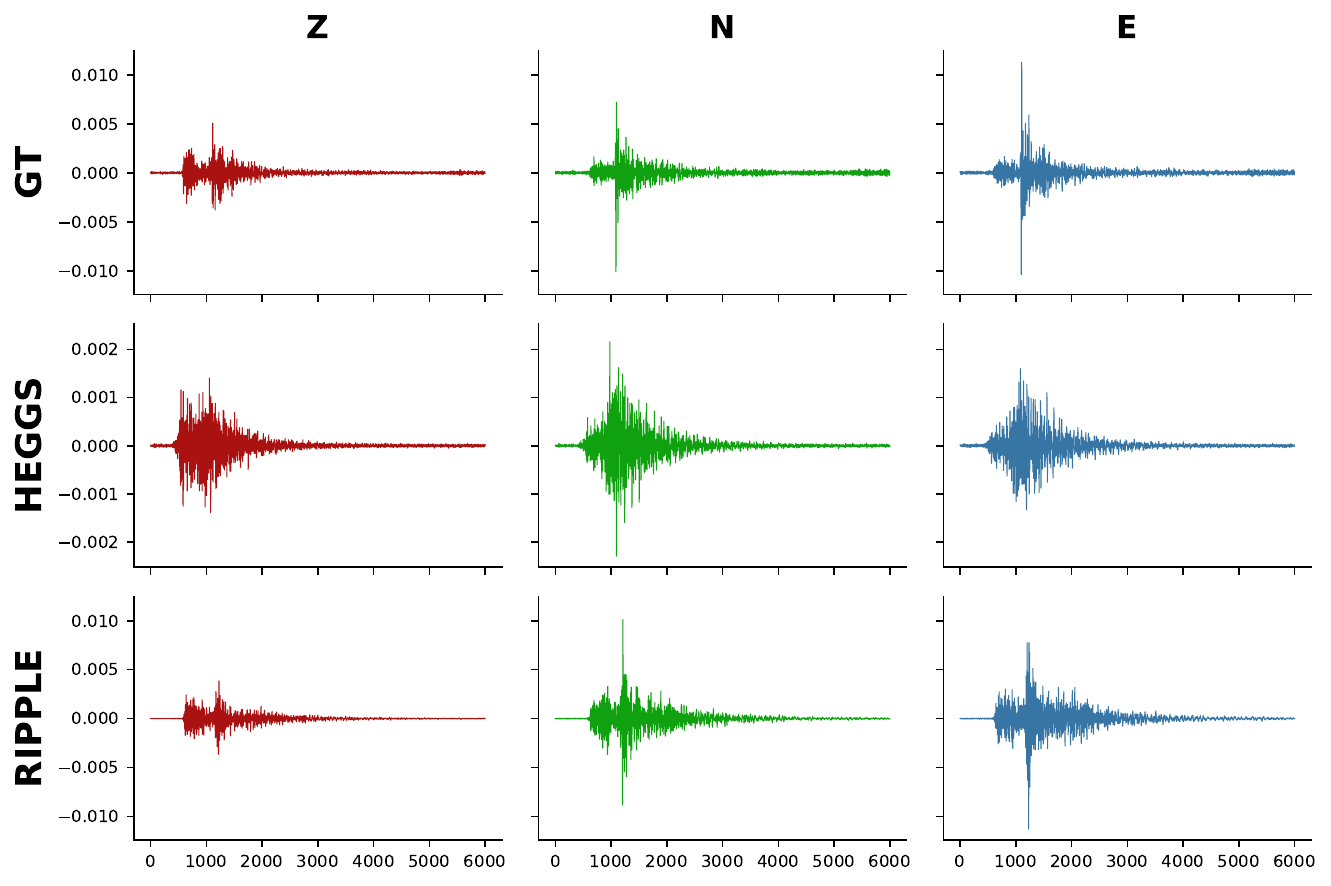}
    \includegraphics[width=0.6\textwidth]{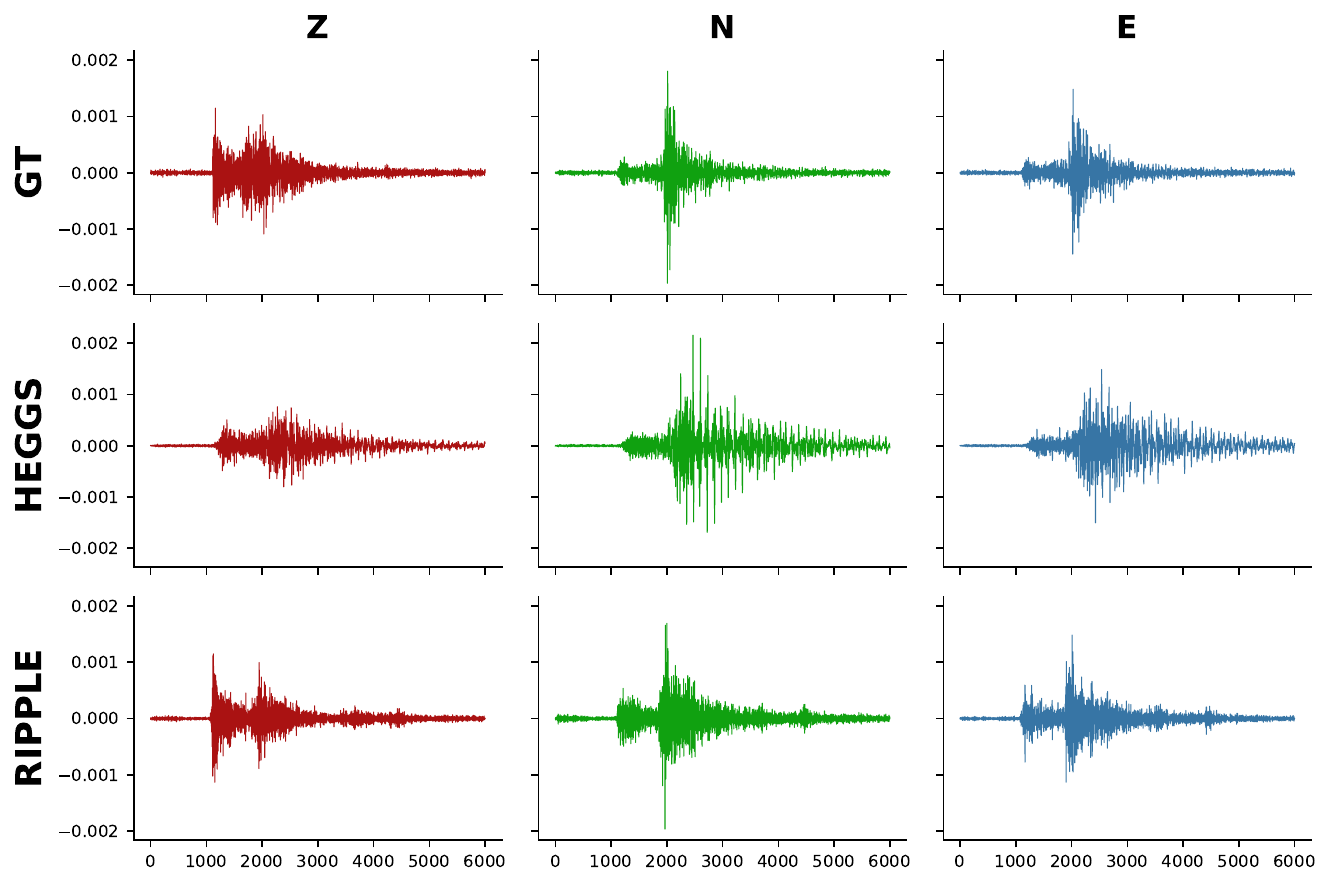}
    \caption{Three-component waveform comparisons for three SCEDC cross-station test samples; amplitude scales are per row.}
    \label{fig:app_seismic_wave}
\end{figure}

Figures~\ref{fig:app_seismic_spec} and~\ref{fig:app_seismic_wave} extend Figure~\ref{fig:seismic_spectrogram} of the main text with additional SCEDC cross-station test samples, in the spectrogram and time domains respectively.

\paragraph{Spectrograms (Figure~\ref{fig:app_seismic_spec}).}
The bottleneck identified in the main text is consistent across events: HEGGS recovers the coarse temporal envelope but attenuates energy above
${\sim}20$\,Hz and blurs the P/S onset structure, an artifact of denoising in a compressed VAE latent. RIPPLE, operating directly on the uncompressed disentangled representation, preserves the broadband texture and the onsets across all three events---the spectral counterpart of the PGA and Arias intensity gaps in Table~\ref{tab:seismic_main}.

\paragraph{Waveforms (Figure~\ref{fig:app_seismic_wave}).}
The same gap is visible in the time domain; note the per-row amplitude scales. HEGGS systematically underestimates peak amplitudes---in the first sample its peaks reach roughly half the target's---which is the waveform-level signature of its PGA deficit in Table~\ref{tab:seismic_main}, while its loss of high-frequency content manifests as overly smooth coda. RIPPLE recovers the peak amplitudes, the sharp arrivals, and the decay of the coda on all three components.

\end{document}